%% file: main.tex
\documentclass[10pt]{iopart}
\usepackage{citesort}
\usepackage{xcolor}
\usepackage{graphicx}
\usepackage{enumerate}
\expandafter\let\csname equation*\endcsname\relax
\expandafter\let\csname endequation*\endcsname\relax
\usepackage{amsmath}
\usepackage{multirow}
\usepackage{adjustbox}
\usepackage{indentfirst} 
\usepackage{booktabs}
\usepackage{tabularx}
\usepackage{graphicx}
\usepackage{soul,color,xcolor}      
\definecolor{myColor}{rgb}{0.8039,0,0}
\begin{document}
\newcommand{\gguide}{{\it Preparing graphics for IOP Publishing journals}}
\renewcommand{\thefootnote}{\fnsymbol{footnote}}
\setcounter{footnote}{-1}

\title{Multi-scale Quaternion CNN and BiGRU with Cross Self-attention Feature Fusion for Fault Diagnosis of Bearing}

\author{Huanbai Liu$^1$, Fanlong Zhang$^{1,*}$\footnote{* Author to whom any correspondence should be addressed.}, Yin Tan$^1$, Lian Huang$^2$, Yan Li$^3$, Guoheng Huang$^1$, Shenghong Luo$^4$ and An Zeng$^1$}
\address{
$^1$ School of Computer Science and Technology, Guangdong University of Technology, Guangzhou, China. \\
$^2$ Department of Applied Electronics, Guangdong Mechanical and Electrical College, Guangzhou, China. \\
$^3$ School of Artificial Intelligence, Shenzhen Polytechnic University, Shenzhen, china.\\
$^4$ Department of Computer Science, University of Macau, Macao, China.
}

\ead{\mailto{zhangfanlong@gdut.edu.cn (Corresponding author)}}
\vspace{10pt}
\begin{indented}
\item[]January 2024
\end{indented}


\begin{abstract}

\noindent In recent years, deep learning has led to significant advances in bearing fault diagnosis (FD). Most techniques aim to achieve greater accuracy. However, they are sensitive to noise and lack robustness, resulting in insufficient domain adaptation and anti-noise ability. The comparison of studies reveals that giving equal attention to all features does not differentiate their significance. 
In this work, we propose a novel FD model by integrating multi-scale quaternion convolutional neural network (MQCNN), bidirectional gated recurrent unit (BiGRU), and cross self-attention feature fusion (CSAFF). We have developed innovative designs in two modules, namely MQCNN and CSAFF. Firstly, MQCNN applies quaternion convolution to multi-scale architecture for the first time, aiming to extract the rich hidden features of the original signal from multiple scales. Then, the extracted multi-scale information is input into CSAFF for feature fusion, where CSAFF innovatively incorporates cross self-attention mechanism to enhance discriminative interaction representation within features. Finally, BiGRU captures temporal dependencies while a softmax layer is employed for fault classification, achieving accurate FD. To assess the efficacy of our approach, we experiment on three public datasets (CWRU, MFPT, and Ottawa) and compare it with other excellent methods. The results confirm its state-of-the-art, which the average accuracies can achieve up to 99.99\%, 100\%, and 99.21\% on CWRU, MFPT, and Ottawa datasets. Moreover, we perform practical tests and ablation experiments to validate the efficacy and robustness of the proposed approach. Code is available at https://github.com/mubai011/MQCCAF.
\end{abstract}

%
\vspace{1pc}

%
\submitto{Measurement Science and Technology}
%
\maketitle
%
\ioptwocol

\section{Introduction}

\noindent Nowadays, many mechanical equipment and production lines have been built to produce various types of products. However, prolonged operation of equipment is subject to constant disturbances from various external factors, increasing the likelihood of accidents and potentially leading to serious safety incidents. The rolling bearing, as a fundamental component of mechanical equipment, plays a crucial role in load support and transmission. Studies have shown that more than 40\% of mechanical failures are related to the failure of rolling bearings~\cite{10083034}. These failures can significantly affect equipment productivity and lifespan, potentially leading to severe damage, hazardous accidents, and substantial financial losses~\cite{li2022fault}. Hence, it is essential to discover a proficient and intelligent approach for the fault diagnosis (FD) of bearing to avoid production disruptions and enhance productivity~\cite{an2023fault,mushtaq2021deep}.

The recent rise in deep learning technologies has provided us with a fresh perspective for addressing FD and has yielded remarkable achievements.
The primary concept of deep learning for addressing FD is integrating feature extraction and classification within a cohesive framework, eliminating human intervention.
This approach effectively overcomes the limitations of traditional machine learning methods~\cite{shao2021fault}, which require an extensive feature selection process before fault identification.
The prevalent deep learning architectures include deep belief network (DBN)~\cite{tang2018fisher}, convolutional neural network (CNN)~\cite{huang20201dcnn}, and recurrent neural network (RNN)~\cite{zhang2021fault}.
CNN has attracted widespread attention owing to its robust feature extraction capabilities.
Two branches of CNN have been utilized for FD: one-dimensional CNN (1DCNN) and two-dimensional CNN (2DCNN).
The primary use of 2DCNN is to process 2D images.
To utilize a 2DCNN for FD, Husari \textit{et al.}~\cite{9801859}, Liu \textit{et al.}~\cite{10167183} and He \textit{et al.}~\cite{10240667} chose separate preprocessing methods for one-dimensional original signals to convert them into two-dimensional images, and then used 2DCNN for feature extraction.

Although 2DCNN has demonstrated a strong performance, the process of converting 1D signals into 2D data requires careful consideration.
However, 1DCNN does not necessitate any transformation of raw signals and has emerged as the dominant approach in FD.
Zhang \textit{et al.}~\cite{zhang2017new} proposed a 1DCNN with a wide first-layer kernel (WDCNN) for the FD of raw vibration signals. Wide kernel convolution was developed to extract robust features and reduce high-frequency noise interference.
Based on the one-dimensional quadratic neuron structure, Liao \textit{et al.}~\cite{10076833} developed an interpretable FD model by learning quadratic functions to generate attention maps.
Jiang \textit{et al.}~\cite{jiang2018multiscale} proposed a multi-scale convolutional neural network (MSCNN) to convert raw vibration signals at different scales to improve the representativeness of the features.
Junior \textit{et al.}~\cite{junior2022fault} utilized multi-sensor signals as 1DCNN inputs for the FD of a motor, where each sensor signal is fed into a CNN structure to extract the hidden features and the final features are a fusion of all types of sensors.

While the previously mentioned 1DCNN has achieved satisfactory results, it may ignore features with long-term dependencies~\cite{jiang2018intelligent}.
Therefore, the performance of FD can be improved via RNN, as it involves extracting features that exhibit long-term dependence. 
Lei \textit{et al.}~\cite{lei2019fault} developed an intelligent FD model based on long short-term memory (LSTM).
LSTM is designed to utilize a three-state gate mechanism to extract rich hidden features from multivariate raw timing signals for the FD of wind turbines.
Liu \textit{et al.}~\cite{liu2018fault} combined a nonlinear autoencoder and a gate recurrent unit (GRU) for FD. 
Nacer \textit{et al.}~\cite{nacer2023novel} proposed a bidirectional long and short-term memory network (BiLSTM) to extract features from fast Fourier-transformed 2D images for FD.
Furthermore, some researchers have integrated 1DCNN and RNN architectures to maximize their respective benefits in the context of FD. 
Huang \textit{et al.}~\cite{huang2021fault} developed a deep model based on CNN-RNN, where they employed empirical modal decomposition (EMD) to construct the input of CNN-RNN to reduce the influence of noise on the FD in CWRU. 
Chen \textit{et al.}~\cite{chen2021bearing} combined a multi-scale CNN and LSTM (MSCNN- LSTM) for FD. 
Two different CNNs are adopted to extract high-frequency and low-frequency features. They are fed into the LSTM to mine features with long-term dependencies.
Ren \textit{et al.}~\cite{ren2023cnn} applied CNNs-LSTM for FD and achieved an accuracy rate of more than 99\% in the identification of nuclear power plants, and the accuracy reaches more than 99\%.
Zhang \textit{et al.}~\cite{zhang2022fault} designed a 1DCNN with an attention mechanism to extract the hidden features first, and the extracted features are fed into a BiGRU (DCA-BiGRU) for FD. We give a table enumerating classic deep learning-based FD methods, as shown in table~\ref{tab:-1}.

\input{tabletex/Table_-1}

Although these mainstream methods have achieved satisfactory practical FD results, they have certain limitations. As shown in table~\ref{tab:-1}, many existing models employ a multi-scale architecture during the feature extraction stage, enabling more comprehensive feature information extraction than a single scale. However, this often necessitates additional preprocessing of input signals into different scales or the utilization of a 1DCNN with varying kernel sizes to achieve multi-scale feature extraction. Preprocessing incurs supplementary computational cost. Furthermore, traditional CNN must pay more attention to the inherent correlations among hidden features during feature extraction, resulting in the loss of valuable information. Additionally, after obtaining multi-scale feature information, these models typically only perform simple concatenation without adequately considering the representation of critical region features and internal interactions between different scales, leading to redundant information. These factors lead to the fact that their models still have great potential for improvement in the ability to resist noise interference and domain adaptation. At the same time, their complex network structure also increases the computational burden.

In recent studies on quaternion, researchers have achieved remarkable results by employing quaternion convolutional network (QCNN) for processing speech signals~\cite{9414248}\cite{guizzo2023learning}. Quaternion, as extensions of hypercomplex numbers, enables the exploration and preservation of underlying connections within data information through quaternion convolution. This facilitates the extraction of more comprehensive and inherently interdependent feature information. Bearing vibration signals, akin to speech signals, are one-dimensional temporal signals that exhibit inherent correlations among the signal information. Furthermore, recent endeavors have focused on enhancing data by integrating attention mechanisms and multi-scale approaches. Wang \textit{et al.}~\cite{wang2024intermittent} introduced a transform comprising self-attention mechanisms at different scales to augment the features. Shao \textit{et al.}~\cite{shao2024adaptive} embedded a convolutional block attention module encompassing the channel and spatial attention in each scale branch to obtain rich fine-grained features. Yan \textit{et al.}~\cite{yan2024liconvformer} combines a transform design with a broadcast self-attention mechanism to extract critical features from global multi-scale information. Therefore, we firmly believe that amalgamating quaternion convolution with attention mechanisms holds substantial potential for exploring multi-scale structures in FD.

In this paper, we propose a FD model called multi-scale quaternion convolution and BiGRU with cross self-attention feature fusion (MQCCAF). The main innovations include the MQCNN and CSAFF. Drawing inspiration from quaternion convolution, we introduce a multi-scale quaternion convolution module (MQCNN) to effectively capture features exhibiting inherent dependency correlations, enhancing the model's accuracy and robustness. Expanding upon existing research on attention mechanisms, we develop a cross self-attention feature fusion module (CSAFF) to augment the fusion of multi-scale feature information. Additionally, we employ a hybrid CNN-RNN architecture and leverage BiGRU to extract temporal features.

The main contributions of this work are as follows:
\begin{enumerate}[(1)]
    \item Quaternion convolution has been introduced in multi-scale fault signal feature learning for the first time. The MQCNN module has been designed to extract essential correlation information embedded within the signals through hypercomplex quaternion convolution operations, thereby enhancing the robustness and accuracy of the model.
    \item In contrast to previous methods that employed concatenation for fusing multi-scale information, we propose an innovative cross self-attention feature fusion (CSAFF) module. The CSAFF explores potential complementary information between different scales, enhances key feature representations in relevant regions, and achieves improved fusion of multi-scale features.
    \item By leveraging the synergistic integration of MQCNN, CSAFF, and BiGRU in its design, the proposed MQCCAF demonstrates its capability to effectively extract comprehensive and robust key features across multiple scales from raw signals, thereby achieving precise FD. Our experiments verify the effectiveness of the proposed method on the CWRU~\cite{Smith2015RollingEB}, MFPT~\cite{Lee2016ConvolutionalNN}, and Ottawa~\cite{https://doi.org/10.17632/v43hmbwxpm.2} datasets. With good performance in noisy and loading environments, the proposed method can diagnose faults accurately and timely.
\end{enumerate}


\section{Related work}
\label{sec:back}
\subsection{Deep learning in fault diagnosis}
\noindent The advancement of deep learning has facilitated its extensive application in FD~\cite{su2024application}, with numerous models based on neural network such as CNN and RNN~\cite{zhu2023review}, bypassing the need for expert knowledge required by traditional methods~\cite{zhang2023dual}. These models extract rich features from the original signals for direct classification diagnosis. Su \textit{et al.}~\cite{10354240} proposed a hybrid LSTM-based model to extract spatiotemporal features using stacked CNN and LSTM layers and focused on critical features via a convolutional block attention module (CBAM) after the first CNN layer. Li \textit{et al.}~\cite{10295916} developed a 2DCNN-GRU model for spatial feature extraction via 2DCNN, temporal information capture via GRU, and optimized parameters using the vector mean algorithm. Zhang \textit{et al.}~\cite{zhan2023fault6} proposed a deep separable convolutional residual network-based model for FD, incorporating a multi-head attention mechanism layer for critical information attention. However, these models have limitations in terms of single-scale feature extraction. Many researchers~\cite{zhang2023two}\cite{10241216} have argued that multi-scale architecture-based models can extract more comprehensive feature information. Jiang \textit{et al.}~\cite{jiang2018multiscale} introduced coarse-grained processes to represent different time-scale original signals and used traditional CNN for multi-scale feature information extraction. Wang \textit{et al.}~\cite{10241216} captured different frequency band fault information through multi-scale convolutional layers connected by convolution kernels of varying sizes and introduced CBAM for significant feature attention. Hu \textit{et al.}~\cite{10082598} used two CNN network with different kernel sizes for automatic multi-scale signal feature extraction and combined two-layer LSTM as a classifier for temporal feature acquisition, achieving high classification accuracy. Liao \textit{et al.}~\cite{liao2024multi} introduced the residual structure on a multi-scale traditional CNN architecture and combined it with an Enhanced GRU (EGRU) to improve the model's time feature extraction ability. 

However, these multi-scale architecture-based models have certain shortcomings. Traditional CNNs typically use inadequate extraction information and lack mining relevance hidden features. Most existing methods adopt simple concatenation for multi-scale feature fusion, lack a practical feature region focus, and are susceptible to noise obscuration. In contrast, our MQCCAF introduces quaternion convolution into a multi-scale architecture. It uses its four components to enhance the internal interaction during the convolution process and fully extracts multi-scale interoperable information. We also introduce the cross self-attention mechanism in the feature fusion process at different scales to enhance adequate regional feature attention, suppress noise information representation, and realize the discriminative difference of multi-scale feature fusion.

\subsection{Quaternion convolutional neural network}
\input{figtex/Fig_1}

\noindent 
Quaternion convolutional neural network is suitable for exploring the hidden dependence between inputs. 
Based on the particular form of quaternion, it has advantages in representing and processing more complex data and obtaining rich features. 
Therefore, quaternion neural network have been successfully applied in many fields in recent years, such as semantic segmentation method~\cite{zheng2022quaternion}, endoscopic clinical diagnosis~\cite{Yu2023quaternion}, topological phase classification~\cite{lin2023quaternion}, and random noise suppression of seismic data~\cite{li2023multi}. 
In the above works, quaternion are the expansion of complex numbers in four-dimensional space, with one fundamental part and three imaginary parts, which can represent the spatial rotation with more flexibility. Moreover, various types of data maintain internal dependencies under the Hamilton product during convolution to obtain richer feature representations. Although quaternion is widely used in medicine, natural language processing, image recognition, and other fields, it is rarely used in bearing FD. Because quaternion perform well in natural language processing, we refer to quaternion to represent bearing signals and introduce quaternion in the convolution process to obtain the relationship between noise and signal, signal and signal.

In contrast to the traditional convolution in the multi-scale architecture model applied in the field of FD in recent years~\cite{shao2024adaptive}\cite{yan2024liconvformer}, the convolution process of quaternion convolutional neural network is based on Hamilton multiplication, which can obtain richer potential features in the process of feature extraction.

\subsection{Self-attention mechanism}
 
In recent years, the attention mechanism has performed better in learning the correlation between different parts of feature elements, so it is widely used in feature extraction in various fields. For example, attention mechanisms extract relevant features in an end-to-end deep learning framework to predict the Remaining Useful Life (RUL) of bearings~\cite{wei2023remaining}. In remote sensing building change detection, attention mechanisms effectively extract high- and low-frequency features~\cite{liang2023enhanced}. Furthermore, attention mechanisms are crucial when diagnosing rolling bearing faults with limited samples in accurately capturing channel and position information~\cite{xue2023rolling}. While attention mechanisms have been extensively used for feature extraction modules in FD, their application is less frequent in feature fusion modules. For instance, Xia \textit{et al.}~\cite{xiao2024bayesian} combined Bayesian and variational self-attention mechanisms to enhance generalization capabilities for feature extraction.

Inspired by channel and position information integration through the above attention mechanism fusion techniques, we introduce the cross self-attention mechanism into our feature fusion module to learn discrepancies between extracted features and fuse essential characteristics. Our research demonstrates that this approach represents an advanced methodology.

\section{The proposed method}
\label{sec:meth}
\noindent
To enhance the extraction of bearing information and improve the accuracy of its assessment, the proposed MQCCAF for FD comprises three components: MQCNN, CSAFF, and Classifier.
The framework is depicted in figure~\ref{fig:1}. The MQCNN contains global feature extraction and multi-scale feature extraction. Moreover, the Classifier consists of time-dependent feature extraction and FD.
The bearing signal's input sequence representation is defined in the Input construction. Global feature extraction is employed to derive comprehensive information from the sequential signal. Multi-scale feature extraction involves employing quaternion convolutional structures of varying scales to get richer multi-scale features.
Cross self-attention feature fusion is employed for multi-scale feature fusion, reducing the influence of redundant features.
BiGRU is utilized to extract long-short dependence features, considering the sequence's time periodicity.
FD aims to distinguish between different fault categories.
Each step is explained in detail in the following parts.

\input{figtex/Fig_2}
\subsection{Input construction}
\noindent 
Given a sequence of $N$ raw vibration signals obtained from various sensors, denoted as $\mathrm{Signal}=\{S_1, S_2, S_3,..., S_n,..., S_N\}$, where each $S_n$ is a time series vibration signal represented as $\mathrm{S_n}=\{X_1, X_2, X_3,$\par\noindent$..., X_t,...,X_T\}$, with $T$ time steps (default is 2048) and $X_t$ denotes the vibration signal value at timestamp $t$.
The label for each signal (indicating faulty type) can be derived via expert knowledge.
For instance, in a given scenario, we have a total of $C$ fault categories, which are represented as $\mathrm{Fault}=\{F_1,F_2,F_3,..., F_f,..., F_C\}$.
Each $F_f$ corresponds to the $f$-th fault misclassification.
This study aims to develop an intelligent FD framework using single-sensor data, and the input of our model is expressed as equation~(\ref{eqn:4}):
\begin{equation}
    \mathrm{I}=\{X_1,X_2,X_3,..., X_t,..., X_T, F_f\}
    \label{eqn:4}
\end{equation}
where I represents the input signal of the sensor, each sensor signal $\{X_1, X_2, X_3,..., X_t,..., X_T\}$ corresponds to one label $F_f$  for training the model.

\subsection{Multi-scale quaternion convolution neural network}
\subsubsection{Global feature extraction}
\ 
\newline
\noindent The input signal is a lengthy temporal sequence encompassing various noises resulting from unpredicted changes during the production process.In addition, the standard convolution technique cannot effectively extract the global long-time corrections between these time points. Fortunately, the comprehensive convolution operation has been designed and validated for extracting this kind of global feature~\cite{zhang2017new}. Our work employs a large convolution kernel $(1*64)$ in the wide convolution layer(WideConv). This allows us to capture the global feature representation of the raw signal, taking into account its long-term correlations. The equation~(\ref{eqn:5}) outlines the procedure for global feature extraction:
\begin{equation}
    f_w=\mathrm{WideConv}(nodes,k,s)(\mathrm{I})
    \label{eqn:5}
\end{equation}
where the raw signal input I is directed to the wide convolution operation, to extract its global features $f_w$. In the wide convolution operation, the term $nodes$ refers to the number of hidden nodes, $k$ represents the size of the convolution kernel, and $s$ denotes the stride size (with a default value of 1).

\subsubsection{Multi-scale feature extraction}
\ 
\newline
\noindent Once we acquire the global feature representation of the raw signal, we develop a multi-scale module to extract the raw signal's rich features at scale, which is motivated by MSCNN~\cite{jiang2018multiscale}.
We specifically utilize $i$ scales (with a default value of 3) to extract the scale representations for the convolution operations. The impact of this default value will be explored in Section 4.6.
In contrast to the MSCNN, we utilize the quaternion convolutional neural network (QCNN) for extracting features at different scales, as depicted in figure~\ref{fig:2}.
The architecture consists of a quaternionic convolutional layer (QConv), a normalization layer, an activation layer, and a maximum pooling layer.
The size of the number of nodes in the quaternion convolution layer is 8 (these values will be discussed in Section 4).
At each scale, we employ three QCNNs ($\mathrm{QCN{N_3}}$) for feature extraction.
Every convolution module possesses a convolution scale of $1*i$, as depicted in equation~(\ref{eqn:6}):
\begin{equation}
\left\{ {\begin{array}{*{20}{c}}
{Q{f_1} = \mathrm{QCN{N_3}}\left( {{f_w},1 \times 2} \right)}\\
{Q{f_2} =\mathrm{QCN{N_3}}\left( {{f_w},1 \times 3} \right)}\\
 \vdots \\
{Q{f_i} = \mathrm{QCN{N_3}}\left( {{f_i},1 \times i} \right)}
\end{array}} \right..
\label{eqn:6}
\end{equation}

QCNN integrates quaternion and CNN operations to extract hidden features.All the inputs, weights, parameters, biases, and outputs are quaternion-based.By incorporating QCNN, the real convolution process in CNN is substituted by a method that extracts hidden features by performing convolution operations on the real matrix. The excellent design enables it to extract hidden features with more accuracy.

The input $f_w$ of QCNN corresponds to the output $f_w$ obtained from the preceding layer's global feature extraction, which encompasses N feature maps. It can be represented as a collection of multiple one-dimensional feature vectors, denoted as $\left( {{x_{(1,1)}},{x_{(1,2)}},...,{x_{(n,n)}}} \right)$. The QCNN operation performs convolution in the real-valued space by convolving a filter matrix with a vector of $f_w$, computed using a single quaternion multiplication. To facilitate quaternion multiplication, we partition $f_w$ into four components: $A_{in}$ represents the first part, $B_{in}i$ represents the second part, $C_{in}j$ represents the third part, and $D_{in}k$ represents the fourth part. Consequently, the input $f_w$ can be transformed and described in a quaternionic form as ${f_w} = {A_{in}} + {B_{in}}i + {C_{in}}j + {D_{in}}k$. The quaternion filter matrix is expressed in the general form as ${\omega ^g} = {\omega ^a} + {\omega ^b}i + {\omega ^c}j + {\omega ^d}k$. In the convolution process, the input vector and the filter matrix are computed based on the Hamilton product. The specific convolution process of QConv is shown in equation~(\ref{eqn:7})-(\ref{eqn:8}):

\begin{equation}
\begin{aligned}
{\omega ^g} \otimes {f_w} =& \left[ {\begin{array}{*{20}{c}}
{{\omega ^a}}&{ - {\omega ^b}}&{ - {\omega ^c}}&{ - d}\\
{{\omega ^b}}&{{\omega ^a}}&{ - d}&{{\omega ^c}}\\
{{\omega ^c}}&d&{{\omega ^a}}&{ - {\omega ^b}}\\
d&{ - {\omega ^c}}&{{\omega ^b}}&{{\omega ^a}}
\end{array}} \right] * \left[ {\begin{array}{*{20}{c}}
{{A_{in}}}\\
{{B_{in}}}\\
{{C_{in}}}\\
{{D_{in}}}
\end{array}} \right]\\
=&\left[ {\begin{array}{*{20}{c}}
{{A_{out}}}\\
{{B_{out}}i}\\
{{C_{out}}j}\\
{{D_{out}}k}
\end{array}} \right]
\label{eqn:7}\\
\end{aligned}
\end{equation}

\begin{equation}
\begin{aligned}
{F_k} &= {\omega ^g} \otimes {f_w} + B \\
 &= \left( {\sum\limits_{i = 1} {\sum\limits_{j = 1}^n m }  + {b^a}} \right) + \left( {\sum\limits_{i = 2} {\sum\limits_{j = 1}^n m }  + {b^b}} \right)i\\
&+ \left( {\sum\limits_{i = 3} {\sum\limits_{j = 1}^n m  + {b^c}} } \right)j + \left( {\sum\limits_{i = 4} {\sum\limits_{j = 1}^n m  + {b^d}} } \right)k
\label{eqn:8}
\end{aligned}
\end{equation}
The input vector and filter matrix are convolved in the above process, and multiple parallel quaternion key hidden feature sets are obtained. Where $F_k$ represents the output of the quaternion convolution layer, ${\omega ^g}$ represents the set of weights, $f_w$ represents the set of input vectors, $B$ represents the set of biases, $m$ represents ${x_{i,j}}\otimes{\omega ^g}_{j, i}$, $x_i$ represents the eigenvalues of the input vectors, and ${\omega ^j}$ represents the values of the weights in the filter matrix. The variables $a$, $b$, $c$, $d$, and $g$ are exclusively employed as labels.

Following the QConv operation, the features extracted from the batch were normalized using quaternion normalization (QuatBn)~\cite{gaudet2018deep}, as described in equation~(\ref{eqn:9}):
\begin{equation}
\begin{aligned}
\mathrm{QuatBN\left(F_k\right)}= \partial\left(\mathbf{W} \times\left(F_k-E\left[F_k\right]\right)\right)+\eta
\label{eqn:9}
\end{aligned}
\end{equation}
where $F_k$ represents the output of the convolution unit, $\mathbf{W}$ represents one of the matrices obtained from the Cholesky decomposition of the inverse covariance matrix $\partial$ represents the input change factor, and $\eta$ represents the offset translation factor.

To enhance the nonlinear expressive capability of QCNN, the ReLU activation function is employed to activate the extracted features.
In addition, the max pooling operation (MAP) is employed to decrease the number of feature parameters.
Thus, the entire procedure of QCNN can be condensed into equation~(\ref{eqn:10}): 
\begin{equation}
\begin{aligned}
\mathrm{QCNN\left(F_k\right)}=\mathrm{MAP}\left(\mathrm{ReLu}\left(\mathrm{QuatBn}\left(F_k\right)\right)\right)
\label{eqn:10}
\end{aligned}
\end{equation}

By employing three QCNN layers, we can acquire multi-scale robust features $\{Qf_1, Qf_2,.., Qf_i\}$, which can then be used for further analysis. 

\subsection{Cross self-attention feature fusion}

\noindent 
The fusion features for FD are formed by integrating the obtained multi-scale features.
The conventional feature fusion method employs a single concatenate operation to fuse the multi-scale features. Nevertheless, including multi-scale features may introduce duplicate information, diminishing the FD's performance.
To reduce the impact of redundant information, we design a cross self-attention feature fusion (CSAFF) scheme.
Unlike the self-attention mechanism, which concentrates on the internal relationship within a single dimension, CSAFF prioritizes the correlation between other scales and selects more vital features during the feature fusion procedure.
The process of CSAFF is given in equation~(\ref{eqn:11}):
\begin{equation}
\begin{split}
{F_F}& = \mathrm{CSAFF}(Q{f_1},Q{f_2},...,Q{f_i})\\
&= \left( {\mathrm{Softmax} (\frac{{Q \otimes {K^T}}}{{\sqrt {{d_k}} }})V} \right) \odot \left( {\prod\limits_{j = 1}^i {Q{f_j}} } \right)
\label{eqn:11}
\end{split}
\end{equation}
where $Qf_i$ denotes multi-scale robust features, $d_k$ denotes the dimensions of Q and K, $Qf_i\in R^{s\times d}$, and $s$, $d$ represent the lengths of the sequences and the dimensions of the features, respectively. The self-attention input sequence is learned, and the weights are initialized by training and mapped into three matrices: Q, K, and V, respectively. Note that the level $\beta$ is obtained by dot multiplication($\bigotimes$) of Q and K as in the dot product formula. The normalized attention weight $\gamma$ (attention score) is obtained by multiplying V by this weight to obtain self-attention attention. Finally, the augmented feature map ($r$) is obtained by multiplying the three sequences using $\bigodot$.

The CSAFF selects the most related hidden features as fusion features $F_f$ for FD, hence enhancing the accuracy of FD.

\subsection{Classifier}
\subsubsection{Time dependence feature extraction}
\ 
\newline
\noindent MQCNN has already extracted the most relevant hidden multi-scale fusion features in the previous stage. 
We designed a BiGRU block to mine the features with long-time dependencies $F_l$ from both past and future sides.
Our design consists of a single BiGRU with 8 nodes, specifically created to extract this type of feature, as depicted in equation~(\ref{eqn:12}):
\begin{equation}
F_l=\mathrm{BiGRU}\left(F_f\right)
\label{eqn:12}
\end{equation}
where $F_l$ represents the fusion feature containing long and short-term features, and $F_f$ is the fusion feature.

\subsubsection{Fault diagnosis}
\
\newline
\noindent
The previous step extracted global, multi-scale, long-time dependencies with attention and fusion features $F_l$ for FD.
To further reduce the complexity of the proposed model for FD, we utilize the global average pooling (GAP) operation to compute a single average value $f$ for each BiGRU channel.
It overcomes the shortcomings of the fully connected layer that easily falls into overfitting while significantly reducing the model's complexity.
The procedure can be expressed as: \begin{equation}
f=\mathrm{GAP}\left(F_l\right).
\label{eqn:13}
\end{equation}

Finally, the softmax function is employed to transform the output of the dense layer into probability values ranging from 0 to 1.
These probabilities represent the likelihood of $C$ faulty types.
The predicted fault is determined by selecting the faulty type with the highest probability value, as indicated in equation~(\ref{eqn:14}).
The dense layer consists of node $C$ and transforms the features $F_f$ into a probability vector. Where $F_c$ is the actual value and f is a single average for each BiGRU channel. As stated below:
\begin{equation}
F_c=\mathrm{Softmax} \left(\mathrm{Dense}(f)\right).
\label{eqn:14}
\end{equation}

The above steps describe the forward progress for FD.
The proposed method adopts one gradient descent algorithm with Adam Optimizer to update the parameters of hidden layers.
The loss function employed is cross-entropy, as specified in equation~(\ref{eqn:15}):
\begin{equation}
\mathrm{Loss}=-\frac{1}{N}\ast\sum_{i=1}^{N}\sum_{C=1}^{C}{F_C\ast\log{{\hat{F}}_C}}
\label{eqn:15}
\end{equation}
where ${\hat{F}}_C$ represents the predicted value, and $F_C$ represents the actual value.
In addition, the activation functions utilized in the CNN layers are ReLU.
After training the proposed method, the new coming signal $s_n^\prime$ could be detected timely and accurately using equation~(\ref{eqn:16}): 
\begin{equation}
\text F_{s_n^\prime}= \mathrm{MQCCAF}(s_n^\prime)
\label{eqn:16}
\end{equation}
where $s_n^\prime$ is the new arrival signal and $F_{s_n^\prime}$ is the type obtained by making fault detection on the new signal $s_n^\prime$. Where MQCCAF is our trained model.
Equation~(\ref{eqn:16}) demonstrates that the proposed method is an end-to-end FD framework. This means that it can automatically detect faults by simply providing the input signal.

\section{Experiments}
\label{sec:exp}
\subsection{Datasets}
\noindent To assess the efficacy of our proposed approach, we performed a comparative analysis with other well-established techniques on three specific datasets: CWRU~\cite{Smith2015RollingEB}, Ottawa~\cite{https://doi.org/10.17632/v43hmbwxpm.2}, and MFPT~\cite{Lee2016ConvolutionalNN}.
The detailed description of each subset is described as follows:
\begin{enumerate}[(1)]
\item MFPT Dataset: it was collected by the Society for Mechanical Failure Prevention Technology under two sampling frequencies of 97656 and 48828.
The data includes three healthy states: defects in the outside ring, faults in the inner ring, and the normal state.
Thus, the FD of MFPT is a 3-classification task.
We provide an example to demonstrate these three states, as depicted in figure~\ref{fig:3}.
It indicates that each state exhibits its distinct pattern. The inner ring's faulty state ranges from -20 to 20 during the 2048 time steps, while the other two are from -4 to 4, and the state of normal shocks is more frequent than the other two. 
\input{figtex/Fig_3}
\item CWRU dataset: we use 12k Drive End Bearing Fault Data and Normal Baseline Data from the CWRU dataset. The data was acquired by sampling four distinct diameters (numbered 1, 2, 3, 4) of 0.007", 0.014", 0.021", and 0.028" at a sampling rate of 12 kHz on the drive end. Each load subset comprises three fault states induced by inner race fault (Ir), ball fault (Ba), and outer race fault (Or), exhibiting varying fault diameters. The fault state with a diameter of 0.028" solely arises from inner race and rolling element faults (Ir4, Ba4). Furthermore, each load condition includes a healthy state (He). Consequently, the FD for CWRU data entails a twelve-classification task encompassing Ba1-4, Ir1-4, Or1-3, He.

\item Ottawa Dataset: the vibration signals of rolling bearings in different states under four loads (increasing speed, decreasing speed, decreasing speed after increasing speed, and speed increase after decreasing) are collected by the University of Ottawa, Canada.
The sampling frequency and time were 20000 Hz and 10s, respectively.
It contains 36 small data sets, and the data of rolling bearings are divided into 5 cases, which are normal data (He), inner ring fault data (In), outer ring fault data (Or), ball fault data (Ba), and combined (inner ring, outer ring, and ball) fault data (Co) under different loads. According to the status data of the rolling bearing, it is marked with a label of 1-5 for a 5-classification task. Additionally, the ability of the proposed method to adapt to different domains is validated in Section 4.5 by dividing the data set into four subsets based on the loads: D1, D2, D3, and D4.
\end{enumerate}

\input{tabletex/Table_0}
\input{tabletex/Table_1}
\subsection{Data processing}
\begin{enumerate}[(1)]
\item Running Environment: The experimental batch size is set to 32, the training period to 100, the learning rate to 0.001, the optimizer to Adam, and the early stopping(patience) to 10 for all the different models.
To assess the performance of our method in comparison to other leading methods, we compute the average accuracy(Acc) over ten times.
This will help mitigate the impact of randomness in deep learning models.
In addition, all experiments were set up to be executed on the NVIDIA GeForce RTX 3090, utilizing CUDA 11.3 and PyTorch 1.11.0.
\item Training Sample Construction:
We adopted an overlap algorithm (same as MSCNN) to generate the training samples. These samples were set at 2048 to ensure a fair comparison with other methods, as this value is commonly used in current approaches.
The overlap algorithm generated 32880, 13700, and 12000 samples for CWRU, MFPT, and Ottawa data sets.
Subsequently, we partitioned three datasets into training and testing subsets in a randomized manner, adhering to a ratio of 4:1.
Furthermore, during the training phase, we utilized one early-stop strategy to train the model with patience ten during 100 epochs.
Specifically, we partition the training portion of each dataset into two subsets: the real training set and the validation set, with a ratio of 4:1. Suppose the validation loss fails to reduce within ten iterations. In that case, the training process will terminate, and the model with the minimum validation loss will be chosen to detect the faults on the testing set.
Alternatively, 100 epochs will be required for the training process.
\end{enumerate}

\subsection{Comparative analysis}
\input{figtex/Fig_4}
\input{figtex/Fig_5}
\noindent 
We performed a comparative analysis of our approach and several prominent methods, including WDCNN~\cite{zhang2017new}, MSCNN~\cite{jiang2018multiscale}, CNNs-LSTM~\cite{ren2023cnn}, DCA-BiGRU~\cite{zhang2022fault} {(Framework of 1DCNN, attention mechanism, and BiGRU), and QCNN~\cite{10076833} on three datasets to illustrate its effectiveness on three subsets. For different methods, we perform training and testing strictly according to the parameters given by their papers in the actual test, and the specific structural parameters are shown in table~\ref{tab:0}.
The data of this experiment are shown in table~\ref{tab:1}, where Acc refers to the accuracy of the model on the corresponding data sets, Mean Acc is its average accuracy on the three data sets, and the number of parameters(Params) of the model is also calculated.

MQCCAF demonstrates superior performance. Am-ong the three data sets, it has the highest average accuracy for MFPT (100\%) and Ottawa (99.21\%).
For CWRU data, its performance is slightly inferior to that of DCA-BiGRU~\cite{zhang2022fault}. We compute the average mean value of three data sets for each method, and we determine that the proposed approach is exceptionally efficient, boasting an average accuracy rate of 99.73\%. In addition, our proposed method exhibits significantly higher efficiency than other methods in Ottawa.
Its accuracy is approximately 99.0\%, while the accuracy rates of other methods are below 95\%.
These methods are ranked as MQCCAF$>$ QCNN~\cite{10076833}$>$ CNNs-LSTM~\cite{ren2023cnn}$>$ MSCNN~\cite{jiang2018multiscale}$>$ DCA-BiGRU~\cite{zhang2022fault}$>$ WDCNN~\cite{zhang2017new} in terms of average accuracy on three datasets, respectively. Furthermore, the findings indicate that our approach is the least burdensome configuration, with parameters around $20.55\times10^3$. At the same time, another multi-scale CNN structure, MSCNN, is high up to $87.06\times10^3$, almost four times the proposed method.

We present a confusion matrix for each approach on the CWRU dataset and the Ottawa dataset to illustrate their distinctions, as shown in figures~\ref{fig:4} and~\ref{fig:5}. The data in figure~\ref{fig:4} demonstrates that both the proposed method and DCA-BiGRU achieve a classification accuracy of over 99.99\%. However, it is noteworthy that the parameter count for DCA-BiGRU is almost eight times higher than that of the proposed method (as indicated in table~\ref{tab:1}). Other methods primarily exhibit misclassifications on Ba3 and Ba4 samples. For example, WDCNN incorrectly classifies 986 Ba4 samples as Ba2, Ba3, or Or3, while QCNN misclassifies 27 Ba3 samples as Ba4. This phenomenon may be attributed to the relatively similar amplitudes of these four signal types. Based on the Ottawa dataset, as illustrated in figure~\ref{fig:5}, it is evident that these six methodologies are capable of accurately identifying the presence of faulty bearings with an average accuracy exceeding 83.07\%. However, only the proposed method outperforms the others, with misclassified samples below 100, while the other comparison methods still need to achieve this level of accuracy. E.g., the WDCNN and MSCNN cannot detect the Co and In faults since the WDCNN misclassifies 131 Co samples into In, and MSCNN misclassifies 144 Co samples into In, as shown in figure~\ref{fig:5} (a) and (b). CNNs-LSTM also cannot detect Co and In faults, as shown in figure~\ref{fig:5} (c). DCA-BiGRU cannot detect outer ring fault (Or) and healthy state (He), which can be found in figure~\ref{fig:5} (d). It misclassifies 312 Or samples as He, which is very dangerous since detecting the faults as normal could cause serious safety accidents. QCNN cannot detect the signals between He and OR, Be and Co, as shown in figure~\ref{fig:5} (e). The results validated the efficiency of the proposed approach for FD.

\input{tabletex/Table_2}
\subsection{The effectiveness of anti-noise}
\noindent
In actual operation, the bearing vibration signal is easily disturbed by the background noise generated by other operating equipment, such as electromagnetic interference, mechanical vibration, and so on. The noise may mask the fault signal that makes FD inaccurate. The anti-noise ability of the model is essential for correct FD. Therefore, we conduct experiments to simulate different noise environments to check the anti-noise ability of the proposed MQCCAF. 

We apply varying levels of Gaussian noise ranging from -6 dB to 6 dB for the three datasets. We then compared these datasets using the following models: WDCNN, MSCNN, CNNs-LSTM, DCA-BiGRU, QCNN, and MQCCAF. The density of Gaussian noise is represented by the Signal-to-Noise Ratio (SNR) as depicted in equation~(\ref{eqn:17}):
\begin{equation}
S=10 \lg \frac{P_s}{P_n}
\label{eqn:17}
\end{equation}
where $S$ represents the signal-to-noise ratio, $P_n$ represents the noise effective power, and $P_s$ represents the signal effective power.
The comparison results of each method on different datasets are shown in table~\ref{tab:2}.

Table~\ref{tab:2} shows that only the proposed method always performs well under strong noise (-6 dB) while others do not. For example, WDCNN and CNNs-LSTM receive 97.85\% and 99.68\% accuracy on the noise-free data set of CWRU (see table~\ref{tab:1}). However, they only receive 85.62\% and 90.44\% accuracy under the -6 dB. On the contrary, the proposed MQCCAF still achieves 97.26\% accuracy.

The experimental findings demonstrate that the proposed approach surpasses other methods in most cases. It won 14 out of 15 cases, except that DCA-BiGRU performs slightly better than the proposed method (99.99\%  vs 99.95\%).
In addition, we compute the average accuracy of each approach across various noise environments.
The findings demonstrate that the proposed technique surpasses alternative methods in all three datasets, achieving accuracies of 99.23\%, 99.79\%, and 97.60\% for CWRU, MFPT, and Ottawa, respectively.
Another finding is that the WDCNN exhibits the lowest performance on CWRU and MFPT with accuracies of 94.85\% and 93.47\%.
DCA-BiGRU performs the worst on the Ottawa subset, whose average accuracy is 91.06\%. 
Even under a strong noise environment (-6 D=dB), the proposed MQCCAF identifies faults with high accuracy exceeding 95\%  across three subsets.
The results confirmed that our approach possesses a strong ability to counteract noise.

\input{tabletex/Table_3}
\input{tabletex/Table_4}
\subsection{Performance across different load domains}
\noindent To evaluate the MQCCAF's adaptability to various loads, we devise 12 transfer tasks using the Ottawa data set, as shown in table~\ref{tab:3}.
For task $D1\rightarrow D2$, we train the model using a subset of $D1$ for training and then evaluate its performance on $D2$.
Similarly, for task $D2\rightarrow D1$, we train the model with $D2$ and test on $D1$.
This process continues for subsequent tasks.
Furthermore, we evaluate the efficacy of the proposed approach by contrasting it with the aforementioned prominent techniques, as depicted in table~\ref{tab:4}.

According to the results, the MQCCAF completed 11 out of 12 tasks, except task $D3\rightarrow D1$, where the WDCNN slightly outperformed the proposed method, achieving an average accuracy of 93.33\% compared to 92.09\%.
In addition, the proposed method's performance is higher than 95\% in most tasks except for $D3\rightarrow D1$ and $D4\rightarrow D1$, indicating that the proposed method performs well for cross-domain FD.
On the contrary, the others cannot since their performances are lower than 95\% in most cases. Specially, the proposed method receives 99.0\% around performance for tasks $D1\rightarrow D3$, $D2\rightarrow D3$, and $D4\rightarrow D3$, respectively. Another multi-scale CNN structure, MSCNN, performs badly on most tasks, including $D2\rightarrow D4$, $D3\rightarrow D1$, $D3\rightarrow D2$, $D3\rightarrow D4$, $D4\rightarrow D1$, and $D4\rightarrow D2$.
The analysis above has confirmed that the MQCCAF proposed exhibits a strong ability to transfer learning across different domains for FD.

\input{tabletex/Table_5}
\input{tabletex/Table_6}

\subsection{Multi-scale effect comparison experiment}
\noindent 

It is necessary to explore the impact of scale numbers by utilizing a single multi-scale structure to extract the concealed characteristics from the raw signals.
To validate their effectiveness, we conducted five sub-experiments of varying scales.
For the case of 1 scale, we only use one type of convolution filter $1\times2$ to extract the concealed features; designed $[1\times2,1\times3]$ for two scales; designed $[1\times2,1\times3,1\times4]$ for three scales; designed $[1\times2,1\times3,1\times4,1\times5]$ for four scales; designed\ $[1\times2,1\times3,1\times4,1\times5,1\times6]$ for six scales, respectively.
Table~\ref{tab:5} shows the comparative results for Ottawa data.

The findings indicate a positive correlation between the scale numbers and performance improvement.
It achieves an average accuracy of 95.78\% on scale 1, 98.54\% on scale 2, 99.22\% on scale 3, 99.23\% on scale 4, and 99.38\% on scale 5.
The accuracy increases significantly from scale 1 to scale 3, but there is little difference between scale 3 and scale 5.
Nevertheless, the parameters of the proposed approach will noticeably rise as the scale numbers increase, as illustrated in table~\ref{tab:6}. 
Therefore, the proposed method sets the scale number as 3 in default (2,3,4), whose FD performance is almost the same as scale 5, but the parameters only take $20.55\times{10}^3$.

\subsection{The influence of wide convolution and quaternion convolution}
\noindent

We conducted five sub-experiments using various wide kernel sizes (ranging from 16 to 256) on the Ottawa dataset to explore how the proposed model is affected by the wide convolution operation.
The comparative results in terms of ten-time average accuracy and parameters are given in table~\ref{tab:7}. 
The findings indicate that the performance of the proposed approach increases with the wide convolution kernels from 16 to 256, and it is stable after 64 with an average accuracy of 99.0\%.
It receives 98.53\%, 98.83\%, 99.21\%, 99.03\%, and 99.02\% for filter sizes 16, 32, 64, 128, and 256 in the wide convolution operation. Setting the wide kernel size to 64 could achieve the best performance. Furthermore, the parameters of the proposed approach decrease as the kernels widen, ranging from $14.52\times{10}^3$ to $47.38\times{10}^3$.
\input{tabletex/Table_7}

\input{tabletex/Table_8}

In addition, we design five sub-experiments to explore the hidden node’s influence in the QCNN layer.
The results for different nodes 4, 8, 16, 32, and 64 on data Ottawa are given in table~\ref{tab:8}. 
The results show that MQCCAF’s performance increases with the hidden nodes.
Significantly, it increases from 97.69\% to 99.39\% with nodes 4 to nodes 64. By solely utilizing node 8, the MQCCAF achieves an accuracy of 99.21\%, significantly more significant than the accuracy of alternative prominent methods outlined in table~\ref{tab:8}. Moreover, the parameters of this method occupy a maximum of 20.55 × 103.
The results suggest that the proposed approach is not affected by the choice of hidden nodes in the QCNN, thus reducing the time required for deploying the model on different FD datasets.

Therefore, in order to balance the best possible performance and the smallest possible number of model parameters, the convolution kernel size of the wide kernel convolutional layer of the proposed MQCCAF is set to $64\times1$ (WideConv(64)), and the number of nodes of its quaternion convolutional neural layer is set to 8 (QCNN(8)), as shown in table 1.

\subsection{Ablation study}
\noindent

To explore the effectiveness of each part in the proposed method, we design four sub-experiments on three data sets. Specifically, CNN and QCNN were designed to explore the effectiveness of quaternion; QCNN and MQCNN were designed to explore the advantages of multi-scale features; MQCNN and MQCNN$+$CSAFF were designed to explore the effectiveness of the cross self-attention feature fusion. The results of the ten-time average accuracy for different data sets are shown in table ~\ref{tab:9}.
\input{tabletex/Table_9}

When comparing CNN with QCNN, it is evident that QCNN has enhanced the performance for MFPT from 99.41\% to 99.55\%, for CWRU from 99.94\% to 99.95\%, and for Ottawa from 94.61\% to 95.95\%. 
Furthermore, the average precision showed that QCNN enhanced the average accuracy by 0.5\% across three subsets.
Utilizing a multi-scale feature extraction structure in QCNN has enhanced the average accuracy from 98.48\% to 98.95\%.
Finally, the model's accuracy using the CSAFF structure on MQCCAF is further improved, and the accuracy of the three data sets is more than 99.21\%.
The proposed approach, which effectively combines MQCNN and CSAFF structure, achieves high accuracy in FD.
Specifically, it achieves accuracy rates of 99.99\%, 100\%, and 99.21\% on CWRU, MFPT, and Ottawa datasets, respectively.
The results validate the efficiency of every component in the MQCCAF for FD.

\subsection{Understanding the MQCCAF}
\noindent
The innovation of MQCCAF primarily lies in its modules, MQCNN and CSAFF, which have been effectively validated, as shown in table~\ref{tab:9}. For the first time, MQCNN adopts a multi-scale architecture based on quaternion convolution, enabling the extraction of richer and internally correlated multi-scale features from raw signals. In an innovative approach, CSAFF incorporates a cross self-attention mechanism to fuse and enhance the multi-scale features obtained. The excellent accuracy of FD achieved by MQCCAF is demonstrated in table~\ref{tab:1}, along with its lightweight model parameters. Furthermore, tables~\ref{tab:3} and~\ref{tab:4} effectively validate the robustness of MQCCAF against noise and the transfer learning capability.

MQCCAF is a FD model based on multi-scale architecture. Unlike conventional architectures that employ regular convolutions to extract features from hidden regions at different scales, we utilize quaternion convolutions to extract more informative features. Direct concatenation often leads to information redundancy and increased computational burden in the fusion stage of multi-scale features. To address this issue, we introduced a cross self-attention mechanism to selectively facilitate information interaction among scales and discriminatively thereby enhance feature representations in critical areas, improving model robustness and accuracy.

Therefore, the constructed MQCCAF represents a novel end-to-end effective FD model with solid robustness.

\section{Conclusion}
\label{sec:con}
\noindent 
To address the issues of inadequate feature extraction and limited discrimination in scale feature fusion within existing multi-scale intelligent diagnosis models, we propose an end-to-end hybrid FD model called MQCCAF, which integrates a multi-scale CNN-RNN structure. We introduce a multi-scale quaternion convolution framework (MQCNN) to extract features from raw bearing signals directly, entirely train the quaternion network, and enable it to converge to more robust local minima, thereby capturing richer and inherently correlated multi-scale feature information. Additionally, we design a cross self-attention feature fusion module (CSAFF), which enhances the interaction between multi-scale feature information, achieves discriminative feature fusion, and strengthens critical feature representation in practical scale regions. Finally, traditional BiGRU and GAP were combined as classifiers for fault classification. Experiments on three datasets demonstrated that our model can efficiently diagnose faults using only raw signals. The accuracy rates achieved for the MFPT dataset were 100\%, the Ottawa dataset is 99.21\%, and the CWRU dataset is 99.99\%. Moreover, MQCCAF exhibits remarkable noise suppression and commendable transfer learning abilities suitable for cross-domain FD applications. Internal analysis shows that the choice of hidden nodes in MQCNN does not significantly affect the MQCCAF performance, resulting in significant time savings during implementation on different FD datasets.

The MQCCAF presented in this study had some limitations. First, a single FD still has some computational time overhead (0.16ms), which is acceptable in most cases considering the millisecond processing time. Second, the accuracy of MQCCAF in cross-domain fault classification must be further improved. The MQCCAF FD framework is lightweight, highly precise, and robust. It has high application prospects for real-time FD in the fields of wind turbines~\cite{xie2023fault} and industrial equipment intelligent manufacturing~\cite{yan2023review}. In future research, we will explore more lightweight quaternion frameworks and methods combined with domain adaptation to enhance the performance of the model in cross-domain FD.

\section*{Acknowledgements}
\noindent This work was supported in part by the National Natural Science Foundation under Grant 92267107, the Science and Technology Planning Project of Guangdong under Grant 2021B0101220006, Key Areas Research and Development Program of Guangzhou under Grant 2023B01J0029, Science and technology research in key areas in Foshan under Grant 2020001006832, the Key Area Research and Development Program of Guangdong Province under Grant 2018B010109007 and 2019B010153002, the Science and technology projects of Guangzhou under Grant 202007040006, the Guangdong Provincial Key Laboratory of Cyber-Physical System under Grant 2020B1212060069, and the Research Foundation of Shenzhen Polytechnic University under Grant 6022310014K and 6022312054K‌.

\section*{References}


\clearpage

\appendix

\end{document}

%% file: tabletex/Table_-1.tex
\begin{table*}[h]
\centering
\caption{\centering{Comparison of classic methods.}}
\label{tab:-1}
\adjustbox{width=\linewidth}{
\Huge
\begin{tabular}{llll}
\hline
\textbf{Name}            & \multicolumn{1}{l}{\textbf{Application}}  & \multicolumn{1}{l}{\textbf{Advantage}}  & \multicolumn{1}{l}{\textbf{Disadvantage}} \\ \hline \\
\multirow{3}{*}{WDCNN~\cite{zhang2017new}}     & \multirow{3}{*}{Diagnose bearing fault.} & CNN is used to extract spatial features.  & Lack of multi-scale feature extraction. \\ 
& & Wide kernel convolution is used to suppress noise. & There is a lack of extraction of temporal features.    \\ & & & Limitations of traditional CNN.           \\ \\
\multirow{2}{*}{QCNN~\cite{10076833}}     & \multirow{2}{*}{Diagnose bearing fault.} & Quadratic neuron is used for feature extraction.  & Lack of multi-scale feature extraction. \\ 
& & Generating attention maps enhances interpretability & There is a lack of extraction of temporal features               \\ \\
\multirow{4}{*}{MSCNN~\cite{jiang2018multiscale}}     & \multirow{4}{*}{Diagnosing gearbox fault.} & CNN is used to extract spatial features.  & Lack of feature interaction between different scales. \\ 
& & A multi-scale architecture is used. & There is a lack of extraction of temporal features.         \\ & & & Limitations of traditional CNN.   \\
 & & & Raw data needs to be preprocessed.\\ \\
\multirow{2}{*}{CNNs-LSTM~\cite{ren2023cnn}}     & \multirow{2}{*}{Diagnosing nuclear power fault.} & CNN-RNN is used to extract spatio-temporal features.   & Lack of multi-scale feature extraction.  \\ 
& &  & Limitations of traditional CNN.    \\ \\
\multirow{3}{*}{MSCNN-LSTM~\cite{chen2021bearing}}   & \multirow{3}{*}{Diagnose bearing fault.} & CNN-RNN is used to extract spatio-temporal features. & Raw data needs to be preprocessed. \\ 
 & & A multi-scale architecture is used. & Lack of feature interaction between different scales.     \\ & & & Limitations of traditional CNN.  \\ \\
\multirow{3}{*}{DCA-BiGRU~\cite{zhang2022fault}}     & \multirow{3}{*}{Diagnose bearing fault.} & CNN-RNN is used to extract spatio-temporal features.  & Large number of parameters. \\ 
& & Attention mechanism is introduced to enhance features. & Lack of feature interaction between different scales.         \\ & & & Limitations of traditional CNN.   \\ \\
 \hline
\end{tabular}
}
\end{table*}

%% file: figtex/Fig_1.tex
\begin{figure*}[ht]
\begin{minipage}[b]{1.0\linewidth}
    \includegraphics[width=\linewidth]{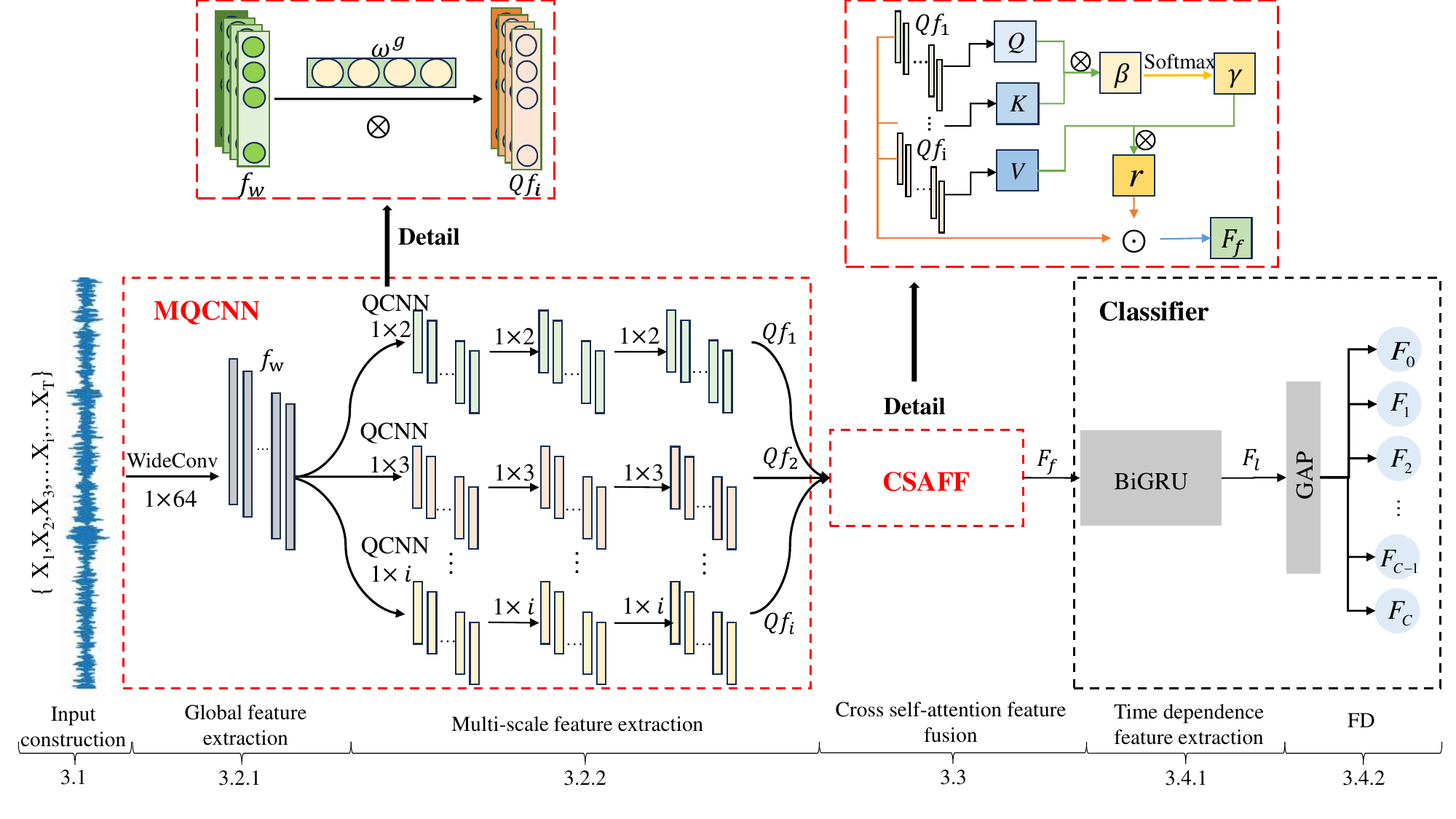}
\end{minipage}
\caption{\centering{Framework diagram of the proposed MQCCAF.}}
\label{fig:1}
\end{figure*}

%% file: figtex/Fig_2.tex
\begin{figure*}[ht]
\begin{minipage}[b]{1.0\linewidth}
    \includegraphics[width=\linewidth]{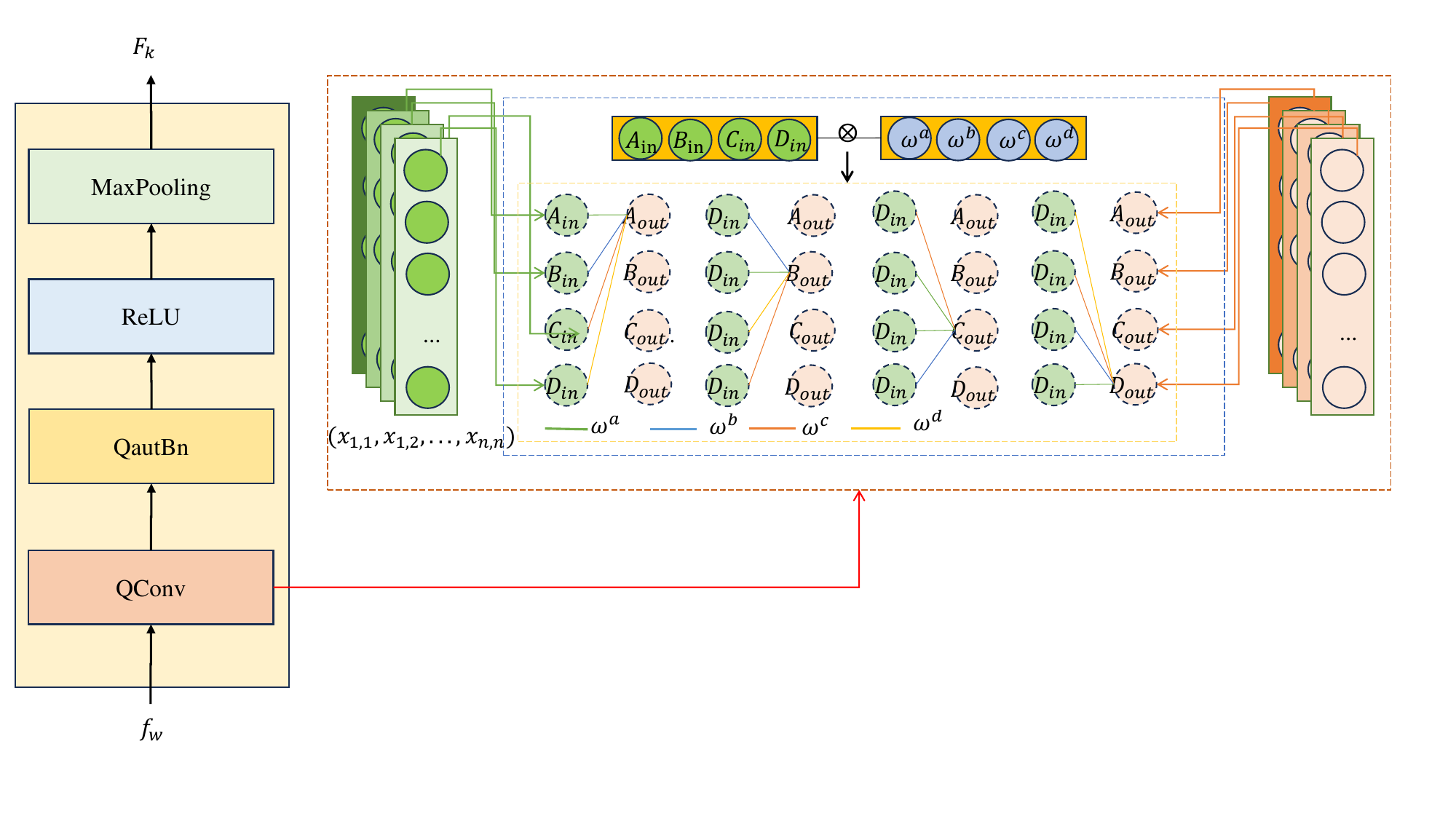}
\end{minipage}
\caption{The flow of feature extraction for the quaternion convolution module(QCNN) is shown in the yellow box, where from bottom to top are the quaternion convolutional layer, quaternion normalization layer, activation layer, and Max pooling layer. The specific implementation of the quaternion convolution layer is shown in the orange box, where green, blue, and orange represent the quaternion components of the input signal, the filtering matrix, and the output of the convolution layer respectively, and the different colored lines in the figure represent the quaternion convolution of different components of the filtering matrix with the input signal to obtain the output vector.}
\label{fig:2}
\end{figure*}

%% file: figtex/Fig_3.tex
\begin{figure}[t]
\begin{minipage}[b]{0.5\linewidth}
    \includegraphics[width=\linewidth]{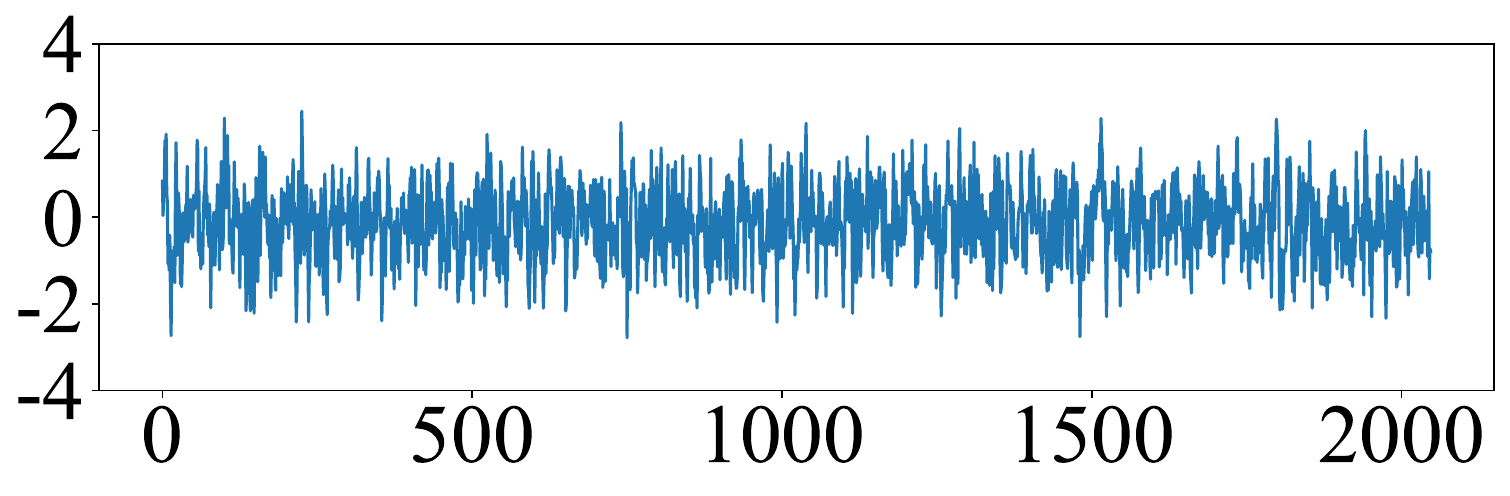}
    \centerline{(a) Normal}\medskip
\end{minipage}
\begin{minipage}[b]{0.5\linewidth}
    \includegraphics[width=\linewidth]{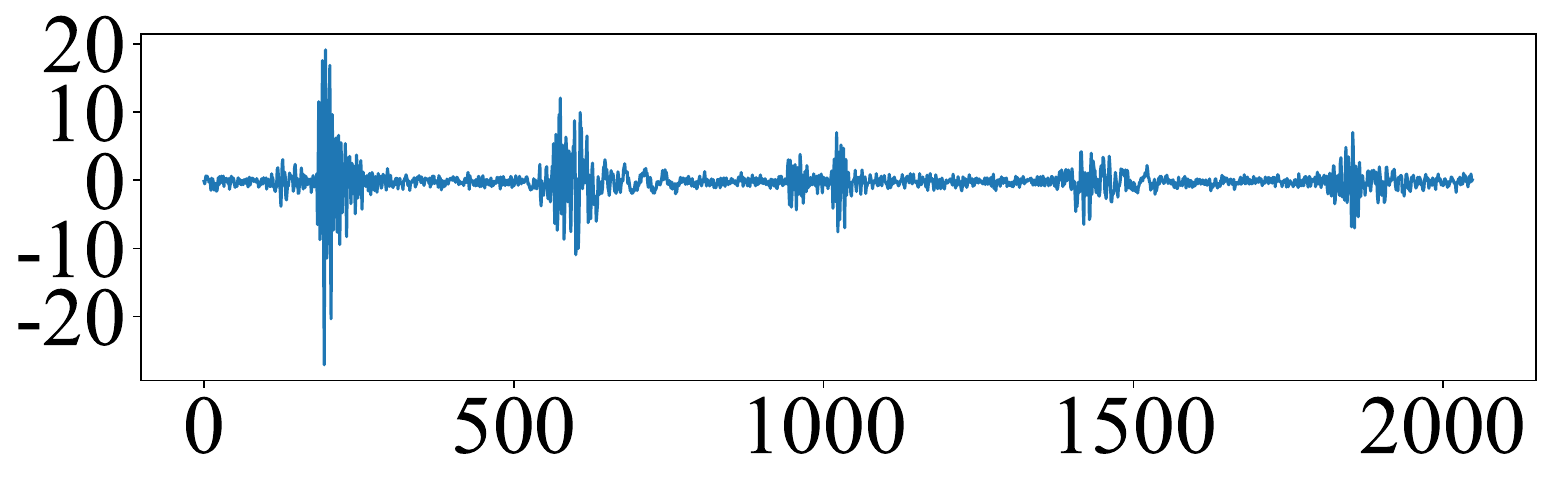}
    \centerline{(b) Inner Ring Fault}\medskip
\end{minipage}
\begin{minipage}[b]{0.5\linewidth}
    \includegraphics[width=\linewidth]{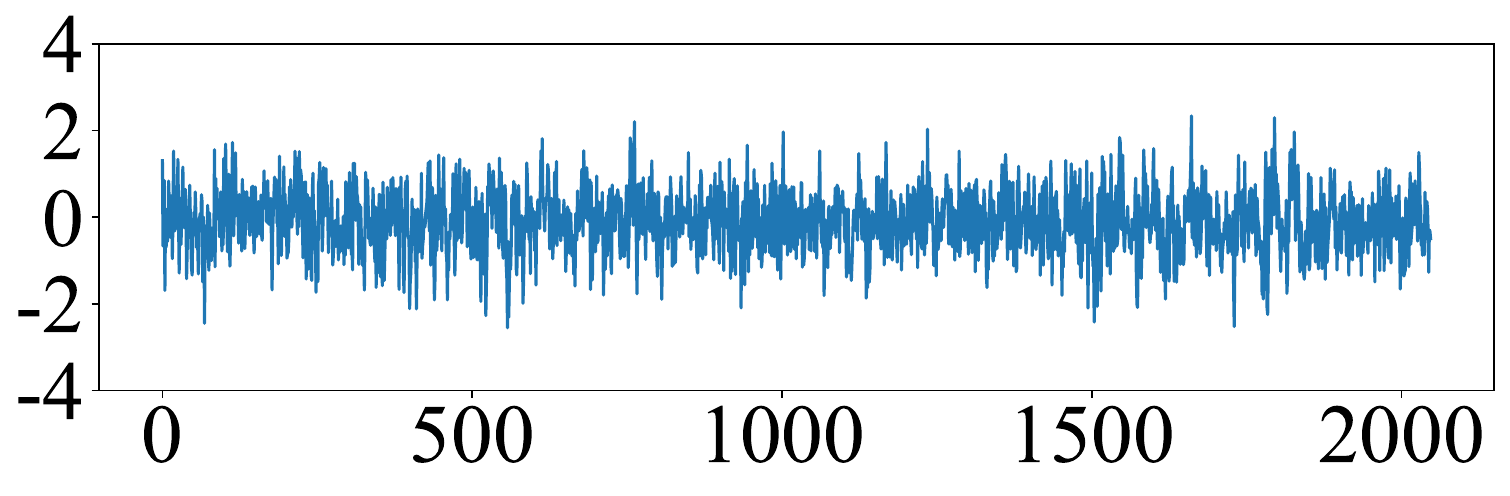}
    \centerline{(c) Outer Ring Fault}\medskip
\end{minipage}
\caption{Raw signals with respect to three classes of MFPT.The horizontal axis represents the time step and the vertical axis represents the signal amplitude.}
\label{fig:3}
\end{figure}

%% file: tabletex/Table_0.tex
\begin{table*}[ht]
\centering
\caption{\centering{Detailed parameter configuration of the proposed MQCCAF with other comparison methods. It should be noted that QCNN of QCNN~\cite{10076833} refers to quadratic convolutional neural network, while QCNN in MQCCAF refers to quaternion convolutional neural network.}}
\label{tab:0}
\adjustbox{width=\linewidth}{
\begin{tabular}{ll}
\hline
\textbf{Method}            & \multicolumn{1}{c}{\textbf{Description}}  \\ \hline \\
\multirow{2}{*}{WDCNN~\cite{zhang2017new}}     & Raw-Conv1D(16)-BN-Maxpool(2)-Conv1D(32)-BN-Maxpool(2)-Conv1D(64)-BN-Maxpool(2)- \\ 
& Conv1D(64)-BN-Maxpool(2)-Conv1D(64)-BN-Maxpool(2)-Dense(100)-Dense(classes)                 \\ \\
\multirow{4}{*}{MSCNN~\cite{jiang2018multiscale}}  & Sub1: Raw-Conv1D(16)-Maxpool(2)-Conv1D(32)-Maxpool(2)-Conv1D(64)-Maxpool(2) \\
& Sub2: Mean(2)-Conv1D(16)-Maxpool(2)-Conv1D(32)-Maxpool(2)-Conv1D(64)-Maxpool(2) \\
& Sub3: Mean(3)-Conv1D(16)-Maxpool(2)-Conv1D(32)-Maxpool(2)-Conv1D(64)-Maxpool(2) \\
& Output: Concatenate(Sub1,Sub2,Sub3)-Dense(100)-Dense(classes) \\ \\
\multirow{1}{*}{CNNs-LSTM~\cite{ren2023cnn}} & Raw-Conv1D(64)-Conv1D(64)-Maxpool(2)-LSTM(16)-Dense(classes) \\ \\
\multirow{3}{*}{DCA-BiGRU~\cite{zhang2022fault}} & Sub1: Conv1D(18)-Maxpool(2) \\
& Sub2: Conv1D(6)-Maxpool(2)-Conv1D(6)-Maxpool(2) \\
& Output: Concatenate(Sub1,Sub2)-Attention(1)-BiGRU-GAp-Dense(classes) \\ \\
\multirow{2}{*}{QCNN~\cite{10076833}} & Raw-QCO(16)-BN-Maxpool(16)-QCO(32)-BN-Maxpool(32)-QCO(64)-BN-Maxpool(64)- \\
&Dense(100)-Dense(classes) \\ \\
\multirow{5}{*}{MQCCAF (ours)} & Raw-WideConv(64)-$f_w$ \\
& Sub1: $f_w$-QCNN(8,2)-QCNN(8,2)-QCNN(8,2) \\
& Sub2: $f_w$-QCNN(8,3)-QCNN(8,3)-QCNN(8,3) \\
& Sub3: $f_w$-QCNN(8,4)-QCNN(8,4)-QCNN(8,4) \\
& Output: CSAFF(Sub1,Sub2,Sub3)-BiGRU(16)-GAP-Dense(classes) 
\\ \hline
\end{tabular}
}
\end{table*}

%% file: tabletex/Table_1.tex
\begin{table*}[t]
\centering
\caption{\centering{The comparative experiment on the effectiveness of classification.}}
\label{tab:1}
\adjustbox{width=\linewidth}
{
\tiny
\begin{tabular}{ccccc}
\hline{MODEL}                     & {DATA} & 
{Acc (\%)}           & {Mean Acc (\%)}                       & Params$(\times10^3)$          \\ \hline
\multirow{3}{*}{WDCNN~\cite{zhang2017new} }    & CWRU          & 97.85$\pm$2.22          & \multirow{3}{*}{83.07$\pm$4.96}          & \multirow{3}{*}{66.99}  \\ 
                                   & MFPT          & 64.73$\pm$3.49          &                                         &                         \\ 
                                   & Ottawa        & 86.64$\pm$9.16          &                                         &                         \\ 
\multirow{3}{*}{MSCNN~\cite{jiang2018multiscale}}    & CWRU          & 99.08$\pm$1.11          & \multirow{3}{*}{96.72$\pm$1.11}          & \multirow{3}{*}{87.07}  \\ 
                                   & MFPT          & 97.80$\pm$0.77          &                                         &                         \\ 
                                   & Ottawa        & 93.29$\pm$1.45          &                                         &                         \\ 
\multirow{3}{*}{CNNs-LSTM~\cite{ren2023cnn}}          & CWRU          & 99.68$\pm$0.09          & \multirow{3}{*}{96.99$\pm$0.41}          & \multirow{3}{*}{32.01}  \\ 
                                   & MFPT          & 98.33$\pm$0.31          &                                         &                         \\ 
                                   & Ottawa        & 92.95$\pm$0.83          &                                         &                         \\ 
\multirow{3}{*}{DCA-BiGRU~\cite{zhang2022fault}}         & CWRU          & \textbf{100.00$\pm$0.00}    & \multirow{3}{*}{95.96$\pm$0.92}          & \multirow{3}{*}{172.98} \\ 
                                   & MFPT          & 100.00$\pm$0.00             &                                         &                         \\ 
                                   & Ottawa        & 87.89$\pm$2.77          &                                         &                         \\ 
\multirow{3}{*}{QCNN~\cite{10076833}}              & CWRU          & 99.91$\pm$0.08          & \multirow{3}{*}{98.43$\pm$0.22}          & \multirow{3}{*}{21.12}  \\ 
                                   & MFPT          & 100.00$\pm$0.00             &                                         &                         \\ 
                                   & Ottawa        & 95.39$\pm$0.57          &                                         &                         \\ 
\multirow{3}{*}{MQCCAF (ours)} & CWRU          & 99.99$\pm$0.02          & \multirow{3}{*}{\textbf{99.73$\pm$0.02}} & \multirow{3}{*}{\textbf{20.55}}  \\ 
                                   & MFPT          & \textbf{100.00$\pm$0.00}    &                                         &                         \\ 
                                   & Ottawa        & \textbf{99.21$\pm$0.05} &                                         &                         \\ \hline
\end{tabular}
}
\end{table*}

%% file: figtex/Fig_4.tex
\begin{figure*}[ht]
\centering
\begin{minipage}[p]{0.35\linewidth}
    \includegraphics[width=\linewidth]{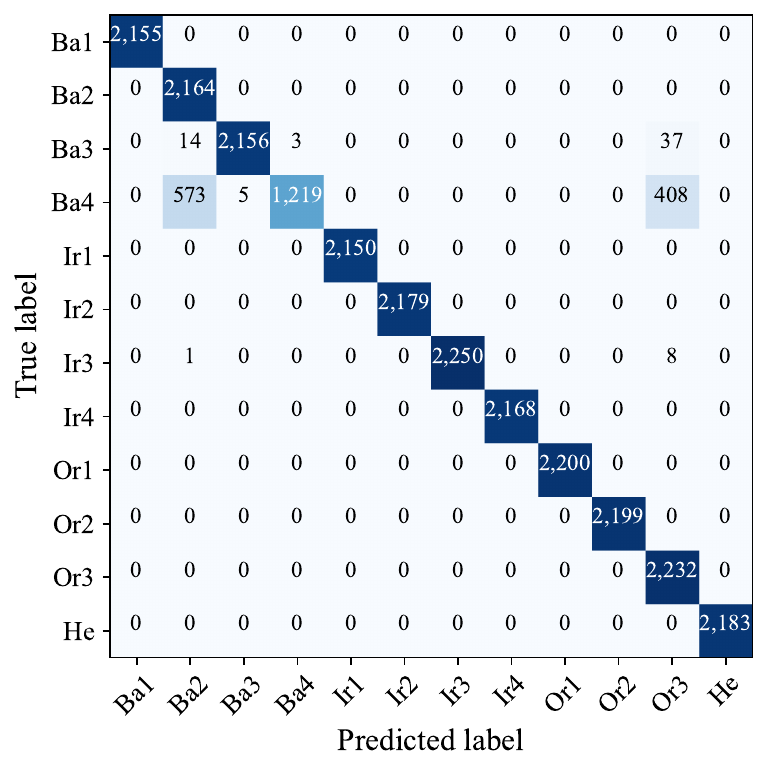}
    \centerline{(a)}\medskip
\end{minipage}
\begin{minipage}[p]{0.35\linewidth}
    \includegraphics[width=\linewidth]{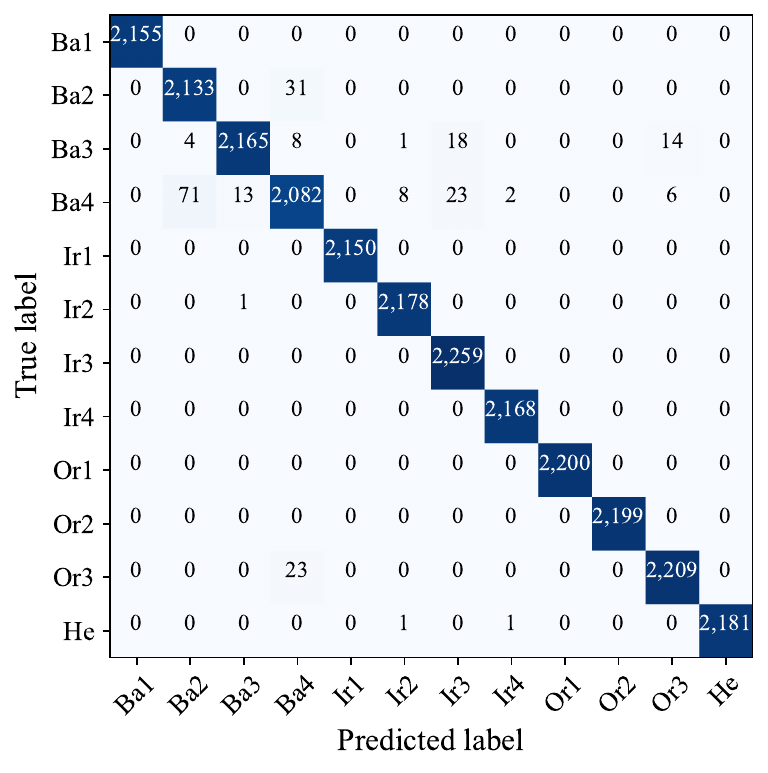}
    \centerline{(b)}\medskip
\end{minipage}
\begin{minipage}[p]{0.35\linewidth}
    \includegraphics[width=\linewidth]{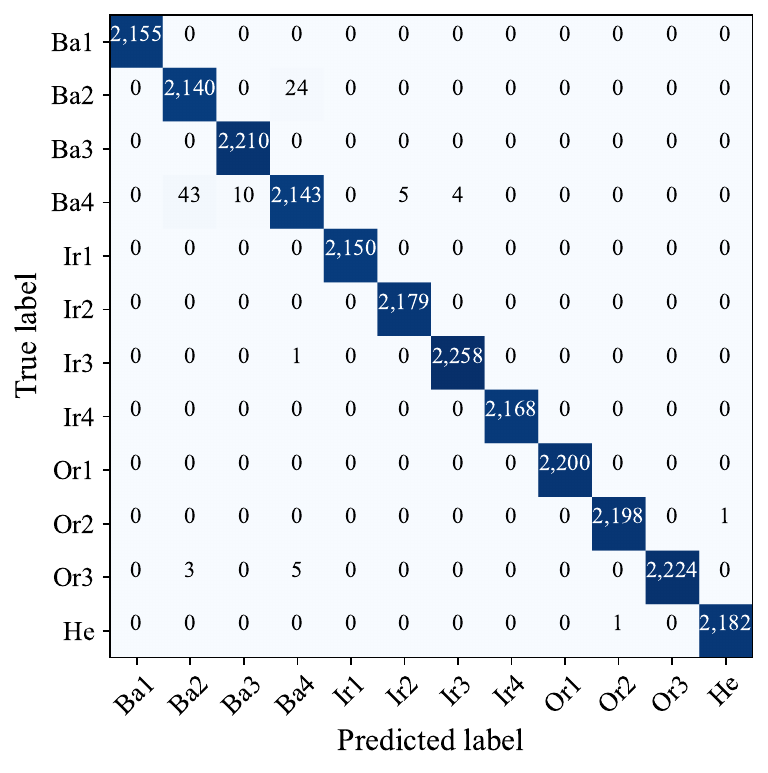}
    \centerline{(c)}\medskip
\end{minipage}
\begin{minipage}[p]{0.35\linewidth}
    \includegraphics[width=\linewidth]{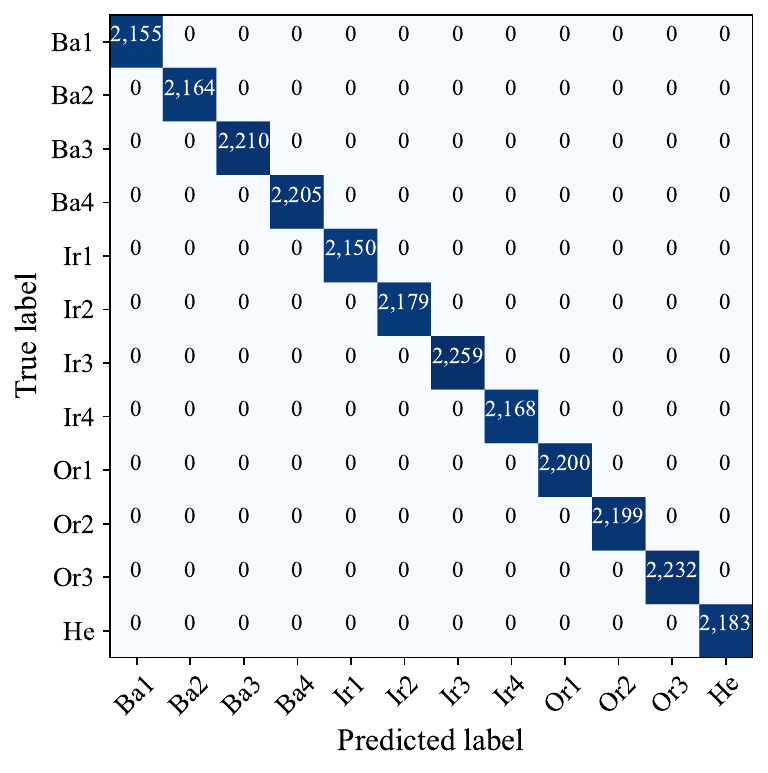}
    \centerline{(d)}\medskip
\end{minipage}
\begin{minipage}[p]{0.35\linewidth}
    \includegraphics[width=\linewidth]{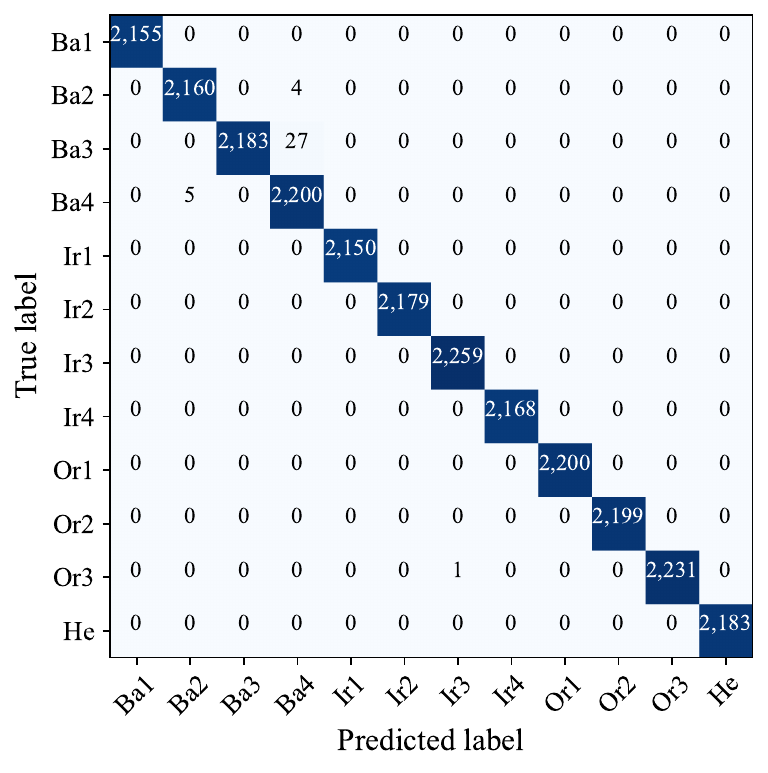}
    \centerline{(e)}\medskip
\end{minipage}
\begin{minipage}[p]{0.35\linewidth}
    \includegraphics[width=\linewidth]{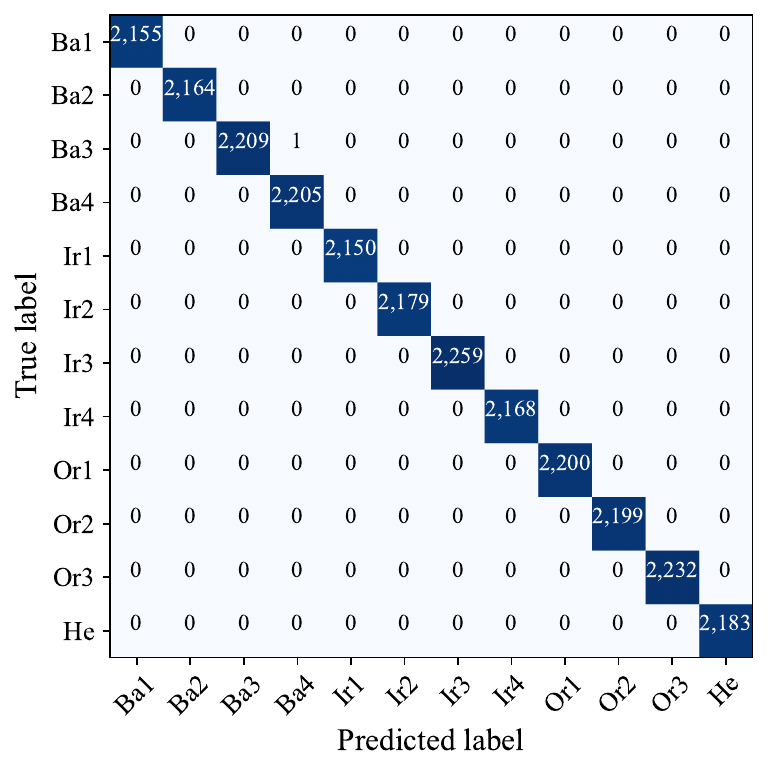}
    \centerline{(f)}\medskip
\end{minipage}

\caption{Confusion matrix effect of different methods on CWRU dataset (a) WDCNN, (b) MSCNN, (c) CNNs-LSTM, (d) DCA-BiGRU, (e) QCNN, (f) MQCCAF (ours).}
\label{fig:4}
\end{figure*}

%% file: figtex/Fig_5.tex
\begin{figure*}[ht]
\centering
\begin{minipage}[p]{0.25\linewidth}
    \includegraphics[width=\linewidth]{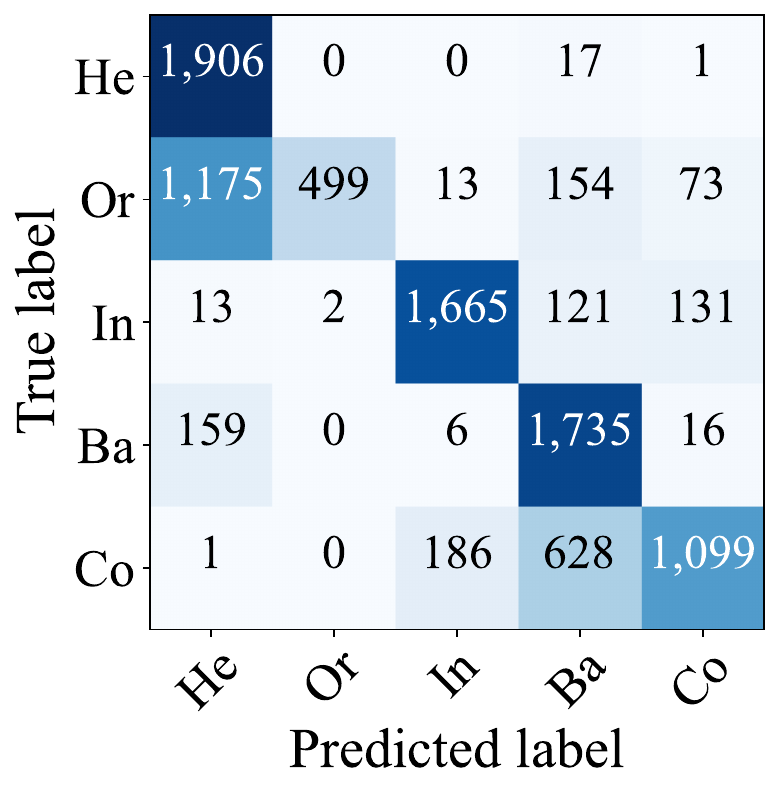}
    \centerline{(a)}\medskip
\end{minipage}
\begin{minipage}[p]{0.25\linewidth}
    \includegraphics[width=\linewidth]{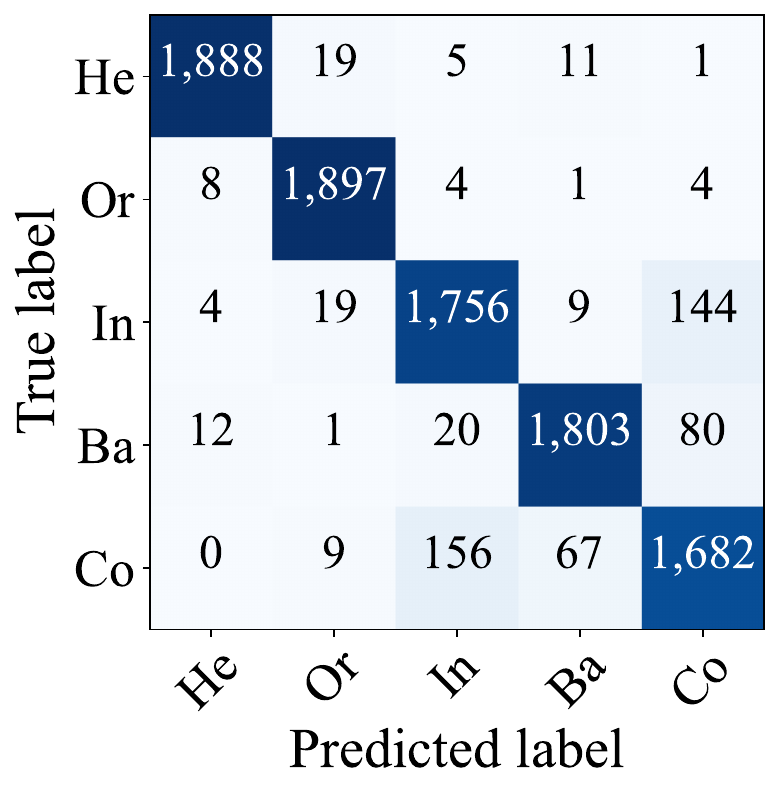}
    \centerline{(b)}\medskip
\end{minipage}
\begin{minipage}[p]{0.25\linewidth}
    \includegraphics[width=\linewidth]{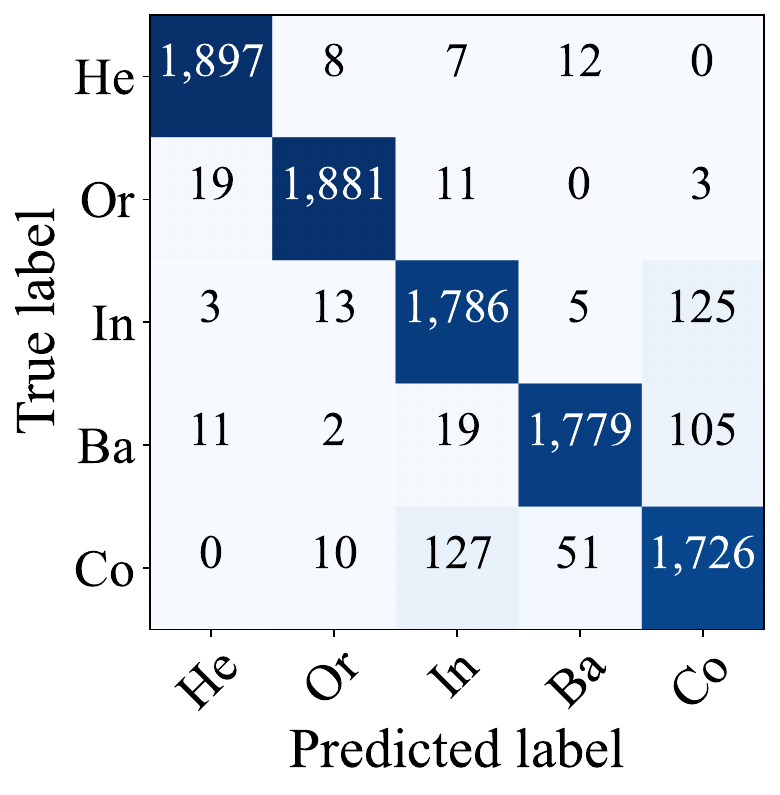}
    \centerline{(c)}\medskip
\end{minipage}
\begin{minipage}[p]{0.25\linewidth}
    \includegraphics[width=\linewidth]{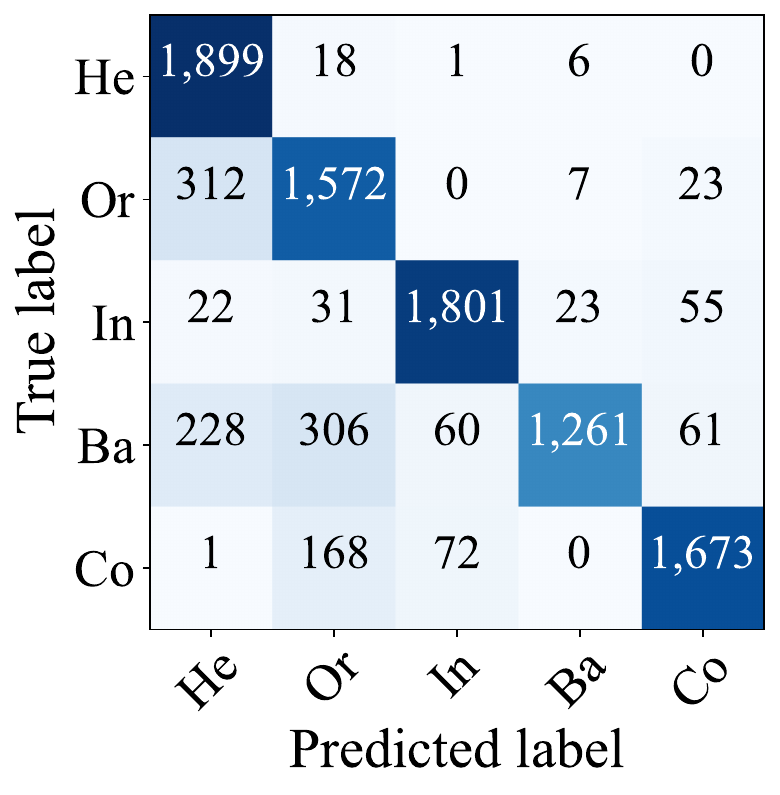}
    \centerline{(d)}\medskip
\end{minipage}
\begin{minipage}[p]{0.25\linewidth}
    \includegraphics[width=\linewidth]{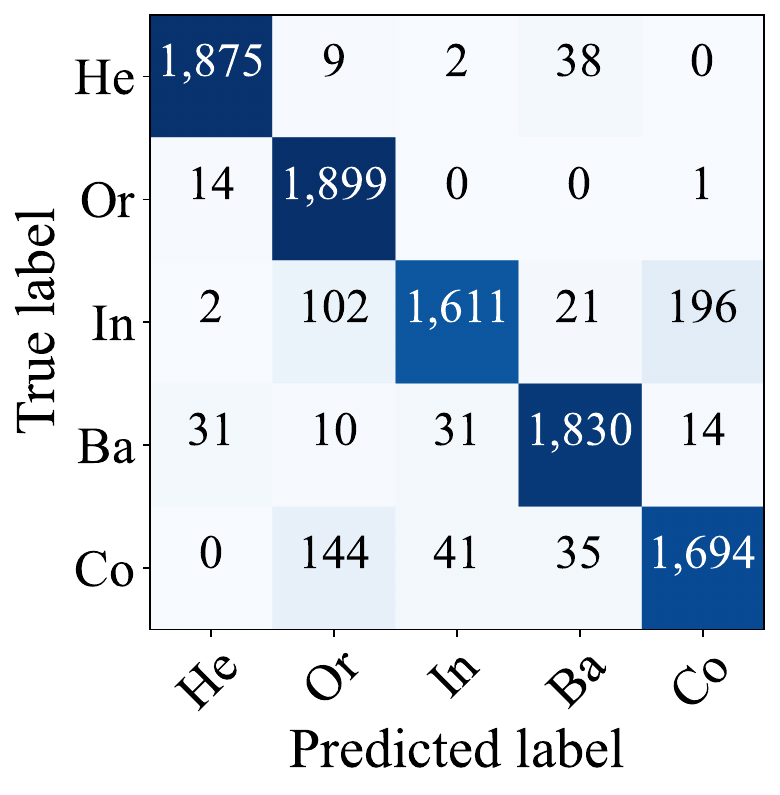}
    \centerline{(e)}\medskip
\end{minipage}
\begin{minipage}[p]{0.25\linewidth}
    \includegraphics[width=\linewidth]{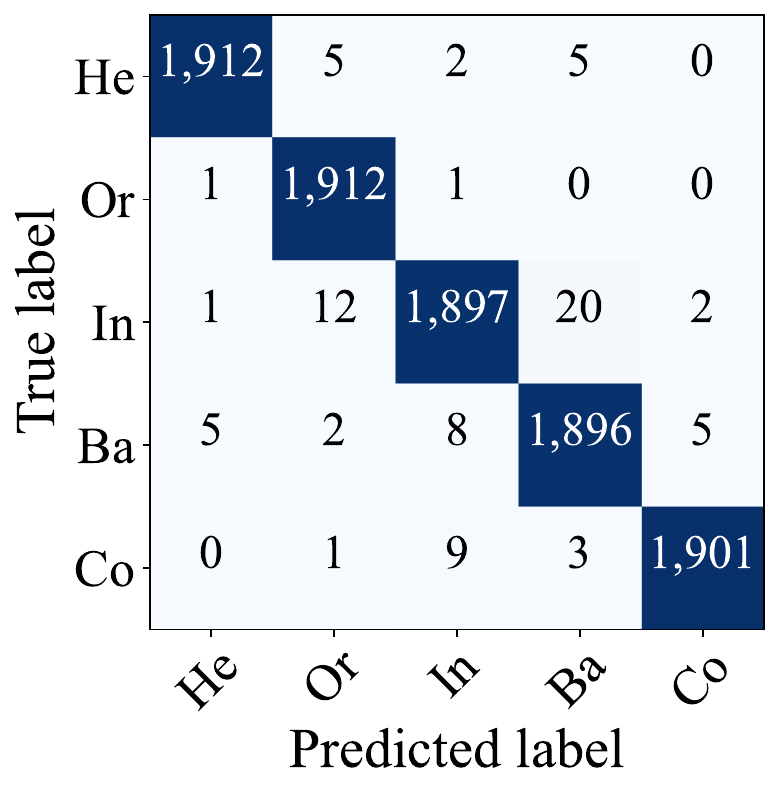}
    \centerline{(f)}\medskip
\end{minipage}

\caption{Confusion matrix effect of different methods on Ottawa dataset (a) WDCNN, (b) MSCNN, (c) CNNs-LSTM, (d) DCA-BiGRU, (e) QCNN, (f) MQCCAF (ours).}
\label{fig:5}
\end{figure*}

%% file: tabletex/Table_2.tex
\begin{table*}[ht]
\caption{\centering{Average results of all compared methods on noisy data. The bold number denotes the best performance denotes the average score of seven noisy conditions.}}
\label{tab:2}
\adjustbox{width=\linewidth}{
\begin{tabular}{cccccccccc}
\hline
& SNR         & -6 (\%)                    & -3 (\%)                    & 0 (\%)                     & 3 (\%)                     & 6 (\%)                     & AVG (\%)           \\ \hline
\multirow{6}{*}{CWRU} & WDCNN~\cite{zhang2017new}       & 85.62$\pm$0.43          & 93.42$\pm$0.55          & 94.97$\pm$0.50          & 97.75$\pm$0.57          & 99.12$\pm$0.25          & 94.58$\pm$0.49          \\ 
                      & MSCNN~\cite{jiang2018multiscale}       & 90.44$\pm$0.53          & 96.19$\pm$0.74          & 97.20$\pm$0.50          & 98.92$\pm$0.29          & 99.50$\pm$0.19          & 96.77$\pm$0.47          \\ 
                      & CNNs-LSTM~\cite{ren2023cnn}   & 92.97$\pm$0.62          & 96.12$\pm$0.53          & 97.72$\pm$0.59          & 99.14$\pm$0.16          & 99.51$\pm$0.19          & 97.25$\pm$0.40          \\ 
                      & DCA-BiGRU~\cite{zhang2022fault}   & 95.14$\pm$0.70          & 97.89$\pm$0.84          & 98.90$\pm$0.39          & 99.55$\pm$0.27          & 99.71$\pm$0.23          & 98.38$\pm$0.51          \\ 
                      & QCNN~\cite{10076833}        & 96.83$\pm$1.35          & 98.86$\pm$0.49          & 99.34$\pm$0.29          & 99.78$\pm$0.13          & 99.82$\pm$0.10          & 99.04$\pm$0.43          \\ 
                      & MQCCAF (ours) & \textbf{97.26$\pm$0.40} & \textbf{99.10$\pm$0.27} & \textbf{99.67$\pm$0.24} & \textbf{99.81$\pm$0.13} & \textbf{99.82$\pm$0.09} & \textbf{99.23$\pm$0.22} \\ 
\midrule
\multirow{6}{*}{MFPT} & WDCNN~\cite{zhang2017new}       & 82.63$\pm$0.69          & 88.14$\pm$0.49          & 96.51$\pm$0.79          & 99.50$\pm$0.43          & 99.91$\pm$0.10          & 93.47$\pm$0.49          \\ 
                      & MSCNN~\cite{jiang2018multiscale}       & 85.23$\pm$0.62          & 93.34$\pm$0.69          & 97.22$\pm$0.47          & 98.77$\pm$0.41          & 99.25$\pm$0.22          & 95.13$\pm$0.50          \\ 
                      & CNNs-LSTM~\cite{ren2023cnn}   & 98.08$\pm$0.17          & 99.13$\pm$0.21          & 99.72$\pm$0.18          & 99.89$\pm$0.05          & 99.92$\pm$0.03          & 99.39$\pm$0.13          \\ 
                      & DCA-BiGRU~\cite{zhang2022fault}   & 98.97$\pm$0.13          & 99.78$\pm$0.08          & \textbf{99.99$\pm$0.04} & 100.00$\pm$0.00          & 100.00$\pm$0.00          & 99.78$\pm$0.05          \\ 
                      & QCNN~\cite{10076833}        & 97.96$\pm$0.22          & 99.46$\pm$0.10          & 99.93$\pm$0.03          & 99.99$\pm$0.00          & 100.00$\pm$0.00          & 99.54$\pm$0.06          \\ 
                      & MQCCAF (ours) & \textbf{99.03$\pm$0.04} & \textbf{99.79$\pm$0.06} & 99.95$\pm$0.02          & \textbf{100.00$\pm$0.00} & \textbf{100.00$\pm$0.00} & \textbf{99.79$\pm$0.03} \\ 
\midrule
\multirow{6}{*}{Ottawa} & WDCNN~\cite{zhang2017new}       & 89.70$\pm$0.90          & 91.78$\pm$0.66          & 93.58$\pm$0.56          & 94.87$\pm$0.54          & 95.27$\pm$0.45          & 93.12$\pm$0.61          \\ 
                      & MSCNN~\cite{jiang2018multiscale}       & 93.95$\pm$0.54          & 95.29$\pm$0.51          & 96.05$\pm$0.61          & 96.95$\pm$0.37          & 96.87$\pm$0.30          & 95.91$\pm$0.46          \\ 
                      & CNNs-LSTM~\cite{ren2023cnn}   & 88.89$\pm$0.24          & 89.99$\pm$0.57          & 90.88$\pm$0.91          & 96.89$\pm$1.67          & 97.92$\pm$0.22          & 93.06$\pm$0.83          \\ 
                      & DCA-BiGRU~\cite{zhang2022fault}   & 82.34$\pm$3.24          & 90.44$\pm$0.76          & 91.42$\pm$0.94          & 93.35$\pm$0.58          & 96.10$\pm$0.33          & 91.06$\pm$1.03          \\ 
                      & QCNN~\cite{10076833}        & 90.58$\pm$1.24          & 93.03$\pm$1.00          & 93.78$\pm$0.52          & 95.78$\pm$0.51          & 0.9653$\pm$0.0039          & 94.07$\pm$0.74          \\ 
                      & MQCCAF (ours) & \textbf{95.02$\pm$0.35} & \textbf{96.80$\pm$0.18} & \textbf{98.24$\pm$0.10} & \textbf{98.67$\pm$0.11} & \textbf{98.99$\pm$0.10} & \textbf{97.60$\pm$0.16} \\ \hline
\end{tabular}
}
\end{table*}

%% file: tabletex/Table_3.tex
\begin{table}[t]
\caption{\centering{Comparison of adaptive capacity of different methods.}}
\label{tab:3}
\adjustbox{width=\columnwidth}{
\begin{tabular}{cccc}
\hline
\multicolumn{1}{c}{\textbf{Source domain}}            & \multicolumn{3}{c}{\textbf{Target domain}}  \\ \hline
\multicolumn{1}{c}{{Train set D1}} & Test Set D2 & \multicolumn{1}{c}{Test Set D3} & Test Set D4 \\ 
\multicolumn{1}{c}{{Train set D2}} & Test Set D1 & \multicolumn{1}{c}{Test Set D3} & Test Set D4 \\ 
\multicolumn{1}{c}{{Train set D3}} & Test Set D1 & \multicolumn{1}{c}{Test Set D2} & Test Set D4 \\ 
\multicolumn{1}{c}{{Train set D4}} & Test Set D1 & \multicolumn{1}{c}{Test Set D2} & Test Set D3 \\ \hline
\end{tabular}
}
\end{table}

%% file: tabletex/Table_4.tex
\begin{table*}[ht]
\centering
\caption{\centering{The comparative experiment on the effectiveness of classification.}}
\tiny
\label{tab:4}
\renewcommand{\arraystretch}{1.5}
\adjustbox{width=\linewidth}
{
\LARGE
\begin{tabular}{ccccccccccccc}
\hline
            & D1-\textgreater{}D2 & D1-\textgreater{}D3 & D1-\textgreater{}D4 & D2-\textgreater{}D1 & D2-\textgreater{}D3 & D2-\textgreater{}D4 & D3-\textgreater{}D1 & D3-\textgreater{}D2 & D3-\textgreater{}D4 & D4-\textgreater{}D1 & D4-\textgreater{}D2 & D4-\textgreater{}D3 \\ \hline
WDCNN~\cite{zhang2017new}       & 92.54\%             & 95.42\%             & 93.54\%             & 93.15\%             & 97.12\%             & 94.06\%             & \textbf{93.33\%}             & 92.53\%             & 94.26\%             & 92.13\%             & 90.82\%             & 97.36\%             \\
MSCNN~\cite{jiang2018multiscale}       & 92.13\%             & 95.27\%             & 90.74\%             & 88.95\%             & 96.20\%             & 89.12\%             & 86.80\%             & 86.95\%             & 89.73\%             & 92.11\%             & 88.91\%             & 95.15\%             \\
CNNs-LSTM~\cite{ren2023cnn}   & 91.25\%             & 95.36\%             & 89.42\%             & 94.15\%             & 97.33\%             & 92.52\%             & 90.19\%             & 89.26\%             & 93.04\%             & 93.12\%             & 88.33\%             & 96.81\%             \\
DCA-BiGRU~\cite{zhang2022fault}   & 93.98\%             & 98.44\%             & 91.65\%             & 95.31\%             & 95.07\%             & 90.67\%             & 87.69\%             & 86.73\%             & 87.83\%             & 92.29\%             & 85.44\%             & 94.58\%             \\
QCNN~\cite{10076833}        & 91.84\%             & 98.66\%             & 95.46\%             & 94.53\%             & 89.74\%             & 94.05\%             & 91.11\%             & 93.07\%             & 95.78\%             & 90.56\%             & 87.14\%             & 96.84\%             \\
MQCCAF (ours) & \textbf{98.53\%}    & \textbf{99.51\%}    & \textbf{97.93\%}    & \textbf{96.92\%}    & \textbf{99.88\%}    & \textbf{97.69\%}    & 92.09\%    & \textbf{95.82\%}    & \textbf{97.88\%}    & \textbf{94.66\%}    & \textbf{95.15\%}    & \textbf{99.53\%} \\ \hline
\end{tabular}

}
\end{table*}

%% file: tabletex/Table_5.tex

\begin{table}[t]
\renewcommand{\arraystretch}{1.5}
\huge
\caption{\centering{Experimental results of multi-scale effect.}}
\label{tab:5}
\adjustbox{width=\columnwidth}{
\begin{tabular}{ccccccc}
\hline
 & {He}             & {Or}                                      & {In}                      & {Bo}        &{Co}         &{AVG}             \\ \hline
MQCCAF & \multirow{2}{*}{98.96\%} & \multicolumn{1}{c}{\multirow{2}{*}{99.48\%}} & \multirow{2}{*}{92.49\%} & \multirow{2}{*}{95.67\%} & \multirow{2}{*}{92.32\%}  & \multirow{2}{*}{95.78\%} \\ 
(1 scales)  &                         & \multicolumn{1}{c}{}                        &                         &                         &                         \\ 
MQCCAF & \multirow{2}{*}{99.53\%} & \multicolumn{1}{c}{\multirow{2}{*}{99.74\%}} & \multirow{2}{*}{97.00\%} & \multirow{2}{*}{98.02\%} & \multirow{2}{*}{98.43\%}  & \multirow{2}{*}{98.54\%} \\ 
(2 scales)  &                         & \multicolumn{1}{c}{}                        &                         &                         &                         \\ 
MQCCAF          & \multirow{2}{*}{99.79\%} & \multicolumn{1}{c}{\multirow{2}{*}{99.84\%}} & \multirow{2}{*}{98.19\%} & \multirow{2}{*}{98.70\%} & \multirow{2}{*}{99.58\%}  & \multirow{2}{*}{99.22\%} \\ 
(3 scales)           &                         & \multicolumn{1}{c}{}                        &                         &                         &                         \\ 
MQCCAF          & \multirow{2}{*}{99.43\%} & \multicolumn{1}{c}{\multirow{2}{*}{99.99\%}} & \multirow{2}{*}{98.29\%} & \multirow{2}{*}{98.85\%} & \multirow{2}{*}{99.69\%}  & \multirow{2}{*}{99.23\%} \\ 
(4 scales)           &                         & \multicolumn{1}{c}{}                        &                         &                         &                         \\ 
MQCCAF          & \multirow{2}{*}{99.74\%} & \multicolumn{1}{c}{\multirow{2}{*}{99.95\%}} & \multirow{2}{*}{98.81\%} & \multirow{2}{*}{98.70\%} & \multirow{2}{*}{99.69\%}  & \multirow{2}{*}{99.38\%} \\ 
(5 scales)           &                         & \multicolumn{1}{c}{}                        &                         &                         &                         \\ \hline
\end{tabular}
}
\end{table}

%% file: tabletex/Table_6.tex
\begin{table}[t]
\renewcommand{\arraystretch}{1.5}
\huge
\caption{\centering{The parameter quantities of MQCCAF at
five scale.}}
\label{tab:6}
\adjustbox{width=\columnwidth}{
\large
\begin{tabular}{cccccc}
\hline
MQCCAF             & 1 scales              & 2 scales              & 3 scales              & 4 scales                            & 5 scales                      \\ \hline
Params$(\times10^3)$               & 13.22                          & 16.67                          & 20.55                          & 28.08                          & 35.11                         \\ \hline
\end{tabular}

}

\end{table}

%% file: tabletex/Table_7.tex
\begin{table}[t]
\centering
\caption{\centering{The effectiveness of wide convolution kernel size in the proposed method.}}
\label{tab:7}
\adjustbox{width=\columnwidth}{
\Huge
\begin{tabular}{cccccc}
\hline
Sizes              & 16              & 32              & 64              & 128                            & 256                      \\ \hline
\multirow{2}{*}{ACC (\%)} & \multirow{2}{*}{98.53$\pm$0.30} & \multirow{2}{*}{98.83$\pm$0.12} & \multirow{2}{*}{99.21$\pm$0.09} & \multirow{2}{*}{99.03$\pm$0.23} & \multirow{2}{*}{99.02$\pm$0.20} \\
                              &                          &                          &                          &                                &                          \\ 
Params$(\times10^3)$                & 14.52                    & 16.71                    & 20.55                    & 29.86                          & 47.38                    \\ \hline
\end{tabular}
}

\end{table}

%% file: tabletex/Table_8.tex
\begin{table}[t]
\renewcommand{\arraystretch}{1.5}
\caption{\centering{The effectiveness of the hidden nodes in the QCNN layer for FD }}
\label{tab:8}
\adjustbox{width=\columnwidth}{
\Huge
\begin{tabular}{cccccc}
\hline
Filters              & 4                     & 8                     & 16                    & 32                             & 64                            \\ \hline
\multirow{2}{*}{ACC (\%)} & \multirow{2}{*}{97.69$\pm$0.68} & \multirow{2}{*}{99.21$\pm$0.09} & \multirow{2}{*}{99.29$\pm$0.03} & \multirow{2}{*}{99.39$\pm$0.05} & \multirow{2}{*}{99.39$\pm$0.30} \\
                              &                                &                                &                                &                                &                               \\ 
Params$(\times10^3)$                & 11.72                          & 20.55                          & 45.71                          & 123.65                          & 390.15                         \\ \hline
\end{tabular}
}
\end{table}

%% file: tabletex/Table_9.tex
\begin{table}[ht]
\renewcommand{\arraystretch}{1.5}
\caption{\centering{Performance of different structures on three datasets.}}
\label{tab:9}
\Huge
\vspace{5pt}
\adjustbox{width=\columnwidth}{
\begin{tabular}{ccccc}
\hline
\multicolumn{1}{c}{\textbf{Model}}     & \multicolumn{1}{c}{CNN (\%)}           & \multicolumn{1}{c}{QCNN (\%)}          & \multicolumn{1}{c}{MQCNN (\%)}         & \multicolumn{1}{c}{MQCNN+CSAFF (\%)}  \\  \hline
MFPT                  & \multicolumn{1}{c}{99.41$\pm$0.18} & \multicolumn{1}{c}{99.55$\pm$0.09} & \multicolumn{1}{c}{99.74$\pm$0.09}    & \textbf{100.00$\pm$0.00}        \\ 
CWRU                 & \multicolumn{1}{c}{99.94$\pm$0.03} & \multicolumn{1}{c}{99.95$\pm$0.04} & \multicolumn{1}{c}{99.97$\pm$0.07}     & \textbf{99.99$\pm$0.02}     \\ 
Ottawa                        & \multicolumn{1}{c}{94.61$\pm$0.38} & \multicolumn{1}{c}{95.95$\pm$0.24} & \multicolumn{1}{c}{97.04$\pm$0.72}     & \textbf{9921$\pm$0.09}     \\ 
Average                        & \multicolumn{1}{c}{97.98$\pm$0.20} & \multicolumn{1}{c}{98.48$\pm$0.12} & \multicolumn{1}{c}{98.95$\pm$0.29}    & \textbf{99.73$\pm$0.04}     \\ \hline
\end{tabular}
}
\end{table}